\documentclass[a4paper,12pt]{amsart} 

\usepackage{amsmath,amsthm,amssymb, amsfonts}
\usepackage{natbib}
\usepackage{geometry}
\usepackage[dvipsnames]{xcolor}
\usepackage{graphicx}
\usepackage{subcaption} 
\usepackage{bm}
\usepackage{algorithm}
\usepackage{algorithmic}

\usepackage{booktabs}
\usepackage{tikz}
\usetikzlibrary{arrows.meta, positioning, fit, calc, backgrounds}
\usepackage{listings} 
\usepackage{mathtools}
\usepackage[T1]{fontenc}
\usepackage{makecell}
\usepackage{comment}

\usepackage{hyperref}
\usepackage{pgfplots}
\pgfplotsset{compat=1.18}
\usepgfplotslibrary{groupplots}

\newcommand{\bmx}{\bm{x}}

\newcommand{\bmz}{\bm{z}}
\newcommand{\bmomega}{\bm{\omega}}
\newcommand{\rme}{\mathrm{e}}
\newcommand{\rmi}{\mathrm{i}}
\newcommand{\bbR}{\mathbb{R}}
\newcommand{\rmd}{\mathrm{d}}

\DeclareMathOperator*{\argmax}{arg\,max}

\begin{document}

\title[]{Cluster-Based Generalized Additive Models Informed by Random Fourier Features} 
\begin{abstract}
In developing data-driven modeling methodologies, there is an ongoing need to reconcile the strong predictive performance of opaque black-box models with the transparency required for critical applications.
This work introduces an interpretable and computationally tractable regression framework for heterogeneous data by combining response-informed spectral representation learning with localized additive modeling. 
The method first fits a random Fourier feature regression model and constructs a spectral feature map from the learned amplitudes and adaptively resampled frequencies, so that the representation reflects predictive variation in the data. 
This representation is then compressed by principal component analysis to obtain a low-dimensional latent embedding, in which a Gaussian mixture model performs soft regime discovery. Within each regime, a cluster-specific generalized additive model captures nonlinear covariate effects through interpretable spline-based univariate smooth functions. The final predictor is formed as a soft mixture of these local additive models, enabling flexible modeling of a nonlinear, heterogeneous structure while preserving interpretability. Numerical experiments across several benchmark regression datasets show that the proposed method consistently improves upon classical globally interpretable baselines while remaining competitive with more flexible black-box models.  
Overall, the framework provides a unified approach to heterogeneous regression that combines predictive adaptivity with interpretable local covariate effects.
\end{abstract}

\author[Huang]{Xin Huang}
\address{Department of Mathematics and Mathematical Statistics, Ume{\aa} University, 901 87 Ume{\aa}, Sweden}
\email{xin.huang@umu.se}

\author[Li]{Jia Li}
\address{Department of Statistics, The Pennsylvania State University, University Park, 16802 PA, USA}
\email{jol2@psu.edu}

\author[Yu]{Jun Yu}
\address{Department of Mathematics and Mathematical Statistics, Ume{\aa} University, 901 87 Ume{\aa}, Sweden} 
\email{jun.yu@umu.se}


\maketitle

\section{Introduction} \label{Section_Intro}
A recurring challenge in modern predictive modeling involves balancing the flexibility required for accurate prediction with the interpretability needed to understand the underlying predictive mechanisms. Highly expressive learners, including deep neural networks (Rumelhart et al. \citeyear{back_propagation_paper}) and ensemble methods (Breiman \citeyear{Random_forest}; Friedman \citeyear{Friedman_pdp}), can approximate complex nonlinear response surfaces with remarkable accuracy, but the mechanisms driving their predictions often remain difficult to interpret. In contrast, classical statistical frameworks such as linear models and generalized additive models (Hastie and Tibshirani \citeyear{GAM_paper}) offer transparent structure and interpretable covariate effects, but this transparency often comes at the cost of reduced flexibility in capturing highly complex or heterogeneous patterns in the data. The resulting tension has motivated two broad directions in explainable machine learning. One focuses on post-hoc explanation of complex predictive models through feature attribution methods, saliency-based analyses, or surrogate models (Simonyan et al. \citeyear{Saliency_maps}; Ribeiro et al. \citeyear{LIME_paper}; Lundberg and Lee \citeyear{SHAP}). The other aims to embed interpretability directly within the model class, so that prediction and explanation arise from the same statistical structure (Breiman \citeyear{Breiman_two_culture}; Doshi-Velez and Kim \citeyear{Towards_interp}).

The tension between model flexibility and interpretability is especially pronounced in heterogeneous regression, where the relationship between covariates and response varies across the input domain (Ma et al. \citeyear{Ma_2020}; Li et al. \citeyear{Li_2022}). In such settings, a single transparent global model may be too rigid to capture regime-dependent structure, whereas highly flexible models may achieve strong predictive performance at the cost of limited interpretability. This challenge motivates locally adaptive approaches that represent a complex regression surface using simpler components defined over different regions of the input space. Mixture-based and locally structured models provide a natural framework for this purpose (McLachlan and Basford \citeyear{Mixture_model}; McLachlan et al. \citeyear{Finite_mixture_model}), introducing latent regimes or soft partitions that approximate heterogeneous global behavior with more interpretable component-wise relationships. However, the effectiveness of these approaches depends critically on how the underlying predictive regimes are constructed. If the induced partition does not align with meaningful predictive variation, the resulting regime-wise interpretable components may fail to provide accurate predictions. These mixture components, therefore, cannot be regarded as reliable explanations.

The partitioning mechanism constitutes a central bottleneck in mixture-based approaches designed to retain interpretability. Classical mixture-of-experts models (Jacobs et al. \citeyear{Adaptive_MOE}; Jordan and Jacobs \citeyear{Hierarchical_MOE}) and related clustering-based strategies (Tang et al. \citeyear{Tang_02}) both derive regimes from the original covariates, either through supervised gating functions or through explicit input space partitioning based on self-organizing maps (Kohonen \citeyear{Kohonen_SOM}). Although these approaches differ in how the partitions are estimated, the resulting regime structure is still largely determined by geometric or distributional organization of the observed covariates, rather than by a representation learned to reflect predictive variation. For regression, however, proximity in the original covariate space does not necessarily imply similarity in the conditional response mechanism.

Recent co-supervised approaches address this limitation by deriving regime information from a trained, high-performing \textit{deep neural network} (DNN), thereby replacing raw geometric partitioning with a prediction-guided latent partition (Seo et al. \citeyear{MLM}; Seo and Li \citeyear{SeoLi2024}).  In these frameworks, the learned regimes are typically used to support simple local predictive models, such as mixture components based on local linear regression. This marks an important advance, since regime formation is no longer tied solely to geometric proximity in the observed covariates. However, the partitioning mechanism is then transferred to a hidden representation generated by a deep architecture that combines auxiliary clustering procedures, making the induced regimes difficult to characterize directly. Seo and Li (\citeyear{SeoLi2024}) partially mitigate this issue by introducing SEE-Net, a DNN architecture designed to produce linearly separable regimes. Nevertheless, their framework is primarily intended to construct surrogate models for deep neural networks, and its extension to other classes of black-box predictors is not immediate. Moreover, because the local components are based on linear regressors, the resulting surrogate models may be too restrictive to capture nonlinear heterogeneous predictive structure.

\subsection{Spectral guidance for locally adaptive regression}

These limitations point to the need for a representation that is sufficiently expressive to reveal heterogeneous response structure while remaining explicit enough to support interpretable regime construction. To this end, we develop a locally adaptive regression framework based on \textit{random Fourier feature} (RFF) regression (Rahimi and Recht \citeyear{Rahimi_Recht}; Rudi and Rosasco \citeyear{Rudi_Rosasco_paper}). Random Fourier features provide explicit finite-dimensional approximations to kernel feature maps, enabling flexible nonlinear regression while preserving the computational tractability of linear models in the transformed feature space (Avron et al. \citeyear{RFF_kernel_paper}; Bach \citeyear{Bach_rff_paper}). By approximating shift-invariant kernel models in an explicit randomized feature space, RFF regression inherits much of the expressive power of kernel methods while avoiding the computational burden of fully nonparametric kernel regression. Consequently, the RFF method provides a suitable foundation for our approach: it can capture nonlinear predictive variation, while its spectral representation remains more structured and transparent than the latent embeddings typical of deep learning architectures.

The spectral form of the RFF model is particularly useful for heterogeneous regression. Fourier components describe variation across different scales and directions in the input space: rapidly varying structure requires higher-frequency components, whereas lower frequencies represent smoother variation. After fitting the RFF model, the learned amplitudes indicate which spectral components contribute most strongly to the prediction. In our framework, these amplitudes are further combined with adaptive frequency resampling based on predictive relevance (Huang et al. \citeyear{RFF_resampling}), yielding a response-informed spectral representation. Because this representation is constructed from the fitted predictive model, it provides a basis for regime discovery that is more closely tied to response variation than to geometric proximity in the original covariate space.



Given the spectral representation derived from the trained RFF model, the proposed method proceeds by separating regime discovery from local additive modeling. Principal component analysis compresses this high-dimensional representation into a low-dimensional latent embedding that captures its dominant modes of variation. A Gaussian mixture model is then fitted in the embedding space, giving regime responsibilities for each observation. These responsibilities define the clustering weights used to train regime-specific \textit{generalized additive models} (GAMs) (Hastie and Tibshirani \citeyear{GAM_book}). Each local GAM models covariate effects through interpretable univariate smooth functions. The final predictor aggregates the local GAMs according to the learned regime weights, producing a mixture-of-GAMs architecture that combines prediction-guided regime formation with interpretable local additive components.

\subsection{Main contributions}

Building on the approach described above, the main contributions of this study are summarized as follows.

\par\medskip
\noindent (1) \textit{Response-informed spectral representation for heterogeneous regression.}
We introduce a response-informed spectral feature map for heterogeneous regression, in which latent regimes are constructed from amplitude-weighted random Fourier features learned through the predictive task rather than from the original covariate space. In settings with spatial or spatio-temporal covariates, the resulting spectral representation admits a meaningful frequency-based interpretation.

\par\medskip
\noindent (2) \textit{A principled framework separating representation, partitioning, and local modeling.}
We develop a representation-guided mixture framework that decouples three key tasks in heterogeneous regression: learning predictive structure, identifying latent regimes, and fitting interpretable local models. By performing regime discovery in a low-dimensional embedding derived from the learned spectral features and combining it with cluster-specific generalized additive models, the proposed approach provides a coherent mechanism for capturing nonlinear heterogeneous structure while maintaining interpretability through component-wise additive effects within each regime.

\par\medskip
\noindent (3) \textit{Empirical performance and interpretable local structure.}
Through experiments on multiple benchmark regression datasets, we demonstrate that the proposed framework consistently improves upon classical globally interpretable models while remaining competitive with more flexible black-box methods. In addition to predictive performance, the method reveals interpretable regime-level patterns and component-wise covariate effects, illustrating how response-informed representations can support accurate prediction and provide insights into local model structure.

The remainder of this paper is organized as follows. Section~\ref{section_related_work} reviews related work on mixture-based local modeling and representation-guided regression. Section~\ref{section_method} presents the proposed method, including the response-informed RFF representation, latent regime discovery with Gaussian mixture modeling, cluster-wise GAM fitting, and the integrated prediction pipeline. Section~\ref{Sec_4} reports numerical experiments on six benchmark regression datasets, together with case studies examining spatial and temporal regime structure. Finally, Section~\ref{section_conclusion} summarizes the main findings and discusses directions for future research. For ease of reference, the notation and abbreviations used throughout this work are summarized in Tables~\ref{tab:abbreviations} and~\ref{tab:nomenclature_new} of Appendix~\ref{app:notations}.

\section{Related Work}\label{section_related_work}

The present work is closely related to two methodological directions. The first concerns locally structured and mixture-based models for heterogeneous regression, where complex global behavior is approximated by simpler region-specific predictors. The second addresses representation-guided modeling, in which a learned feature representation is used to support a simpler or more interpretable downstream predictor. Our framework draws on both perspectives while differing in its use of a response-informed spectral representation to guide latent partitioning and cluster-specific additive modeling.

The first relevant strand studies local modeling strategies for heterogeneous regression. As discussed in Section~\ref{Section_Intro}, mixture-based and locally structured models provide a natural way to represent complex response surfaces through simpler component-wise predictors. Particularly relevant to our work are approaches that emphasize local interpretability, either through structured mixtures or through simplified region-specific predictors (Bastani et al. \citeyear{Bastani_paper}; Ismail et al. \citeyear{Ismail_2022}). Our framework shares this objective, but differs in that the local regimes are not constructed directly from the original covariate space and the component models are generalized additive rather than linear.

The second related strand explores the use of learned representations to support downstream modeling. This perspective appears, for example, in knowledge distillation (Hinton et al. \citeyear{Distilling_Knowledge}), where the predictive structure learned by a complex model is transferred to a simpler one. More broadly, several recent works investigate how random Fourier feature representations can support interpretability, clustering, or adaptive kernel learning. The G-NAMRFF model (Reddy et al. \citeyear{Interpretable_Graph_NN_with_RFF}) incorporates RFF-based kernel approximations into additive models for graph-structured prediction. Although it also emphasizes interpretability, it is formulated for graph prediction and employs a single global additive model rather than cluster-specific local models. Random Fourier features have also been used in clustering-oriented settings. Chitta et al. (\citeyear{Clustering_Induced_Kernel_Learning}) show that clustering in the transformed feature space can efficiently approximate kernel \(k\)-means, while Nguyen et al. (\citeyear{Efficient_Kernel_Clustering_Using_RFF}) combine clustering and kernel learning through reparameterized RFFs in a Dirichlet-process framework. End-to-end kernel learning with generative random Fourier features has also been studied by Fang et al. (\citeyear{Kernel_Learning_Genrative_RFF}), in which the spectral distribution is jointly optimized with a classifier. In contrast to these approaches, our goal is not unsupervised clustering or classification-oriented kernel learning in isolation, but the construction of a supervised spectral representation for response-guided partitioning in heterogeneous regression.

A particularly close line of work investigates local modeling guided by learned latent representations. The mixture-of-linear-models framework and the connected DNN co-supervised architecture (Seo et al.~\citeyear{MLM}; Seo and Li~\citeyear{SeoLi2024}) show how representations learned by deep networks can be used to define predictive regions before fitting simpler local models. Our framework is similar in spirit in that regime construction is guided by a learned representation rather than by raw input geometry. The main differences are that we replace DNN-based latent features with an explicit RFF-based spectral representation and replace local linear predictors with cluster-specific generalized additive models, thereby allowing nonlinear but still interpretable effects within each regime. In addition, for settings with spatial or temporal covariates, the spectral representation itself may admit a meaningful frequency-based interpretation.

Taken together, these lines of work motivate a framework for heterogeneous regression in which a learned predictive representation guides regime construction while local response behavior remains interpretable.

\section{Method}
\label{section_method}

\subsection{Overview and design principles of the proposed framework}

Suppose that we observe a dataset $\{(\bmx_i,y_i)\}_{i=1}^N$ consisting of $N$ independent input-output pairs drawn from a joint distribution of random variables $(X,Y)$, where $X\in\mathbb{R}^p$ denotes the covariate vector and $Y\in\mathbb{R}$ is the associated response. In the regression setting, our objective is to estimate the target function
\begin{equation}\label{eq:m_x_def}
m(\bmx) = \mathbb{E}[Y \mid X=\bmx]\,,
\end{equation}
that is, the conditional expectation of the response given the input $\bmx$.

In heterogeneous regression, accurate prediction requires more than a flexible global approximation of $m(\bmx)$. One also needs to identify regimes in which the conditional response mechanism behaves coherently. Partitions formed directly in the original covariate space may fail to capture this structure, since geometric proximity of covariates does not necessarily imply similarity of predictive behavior. The proposed framework therefore constructs regimes in a representation space informed by the response, rather than in the raw input space. 

Our proposed method separates three tasks that are often conflated in locally adaptive regression:
\begin{enumerate}
    \item \textit{Learning predictive structure.}
    A random Fourier feature regression model provides a global spectral approximation of the regression function. Its learned amplitudes and adaptively resampled frequencies define a response-informed feature map.

    \item \textit{Identifying latent regimes.}
    Principal component analysis compresses the learned spectral representation into a low-dimensional latent embedding. A Gaussian mixture model is then fitted in this embedding, yielding posterior regime probabilities.

    \item \textit{Fitting interpretable local models.}
    Within each latent regime, a GAM captures nonlinear covariate effects through univariate smooth functions.
\end{enumerate}

At a high level, the resulting predictor has a soft mixture form
\[
\tilde m(\bmx)
=
\sum_{\ell=1}^L
w_\ell(\bmx)\,\tilde f^{(\ell)}(\bmx),
\qquad
w_\ell(\bmx)\ge 0,
\qquad
\sum_{\ell=1}^L w_\ell(\bmx)=1\,,
\]
where $\tilde f^{(\ell)}$ is the local GAM associated with regime $\ell$, and $w_\ell(\bmx)$ is the input-dependent regime weight obtained from the latent-space mixture model. 
The following subsections make this construction precise by defining the spectral representation, the latent embedding, the regime weights, and the local additive models.

\subsection{Response-informed latent regime discovery}
\label{sec:response_informed_regime_discovery}

This subsection constructs the latent regimes used by the proposed model. The key idea is first to approximate the regression structure using a response-informed RFF representation, then to project this representation onto a stable, low-dimensional latent space, and finally to fit a probabilistic mixture model in that space.

\subsubsection{Spectral representation via random Fourier features}
\label{sec:rff_representation}

We begin with kernel regression as a motivating nonlinear model for the target function in \eqref{eq:m_x_def}. Given training data $\{(\bmx_i,y_i)\}_{i=1}^N$, a kernel estimator has the form
\begin{equation}\label{eq:m_kappa_func}
m_{\kappa}(\bmx)
=
\sum_{i=1}^N \eta_i\,\kappa(\bmx,\bmx_i)\,,
\end{equation}
where $\kappa:\bbR^p\times\bbR^p\to\bbR$ is a symmetric positive semidefinite kernel and $\bm{\eta}=[\eta_1,\dots,\eta_N]^\top$ is the vector of model coefficients. A common choice for the kernel is the Gaussian \textit{radial basis function} (RBF)
\begin{equation}\label{eq:Gaussian_RBF}
\kappa_\sigma(\bmx,\bmx')
=
\exp\!\left(
-\frac{\|\bmx-\bmx'\|^2}{2\sigma^2}
\right),
\end{equation}
with bandwidth parameter $\sigma>0$. The model coefficients $\bm{\eta}$ can be estimated by kernel ridge regression, equivalently through the penalized least-squares problem
\[
\min_{\bm{\eta}\in\bbR^N}
\left\{
\|\bm{\Xi}\bm{\eta}-\bm{y}\|_2^2+\lambda_\kappa\,\bm{\eta}^\top\bm{\Xi}\bm{\eta}
\right\},
\]
which leads to
\begin{equation}\label{eq:kernel_ridge_regression_gold}
(\bm{\Xi}+\lambda_\kappa\bm{I})\bm{\eta}=\bm{y}\,,
\end{equation}
where $\bm{y}=[y_1,\dots,y_N]^\top$ is the response data vector, $\lambda_\kappa>0$ is a Tikhonov regularization parameter, and $\bm{\Xi}\in\bbR^{N\times N}$ is the kernel matrix with entries $\bm{\Xi}_{ij}=\kappa(\bmx_i,\bmx_j)$. Since $\bm{\Xi}$ is typically dense, solving \eqref{eq:kernel_ridge_regression_gold} by direct methods such as Cholesky factorization requires $\mathcal O(N^3)$ operations, which becomes prohibitive for large $N$. Random Fourier features provide an explicit finite-dimensional approximation to this class of kernel models, avoiding the need to work directly with the full kernel matrix.

The construction is based on \textit{Bochner's theorem} (Wendland \citeyear{SDA_book}; Bach \citeyear{Bach_book}) for shift-invariant kernels. Suppose that $\kappa(\bmx,\bmx')=\kappa(\bmx-\bmx')$ is a continuous positive definite shift-invariant kernel. In the normalized case, and assuming that the associated spectral measure admits a density $\rho:\bbR^p\to\bbR^+$, Bochner's theorem gives the representation
\begin{equation}\label{eq:Bochner_theorem}
\kappa(\bmx-\bmx')
=
\int_{\bbR^p}
\rho(\bmomega)
\rme^{\rmi\bmomega\cdot(\bmx-\bmx')}
\,\rmd\bmomega
=
\mathbb E_{\rho(\bmomega)}
\left[
\zeta_{\bmomega}(\bmx)
\zeta_{\bmomega}(\bmx')^\ast
\right],
\end{equation}
where $\rho(\bmomega)$ is a nonnegative spectral density, $\zeta_{\bmomega}(\bmx)
:=
\rme^{\rmi\bmomega\cdot\bmx}$,
and $(\cdot)^\ast$ denotes complex conjugation. Approximating this expectation by Monte Carlo sampling with frequencies $\{\bmomega_k\}_{k=1}^K$ drawn independently from $\rho$ yields
\begin{equation}\label{eq:MC_approx_kernel}
\kappa(\bmx-\bmx')\simeq
\frac{1}{K}\sum_{k=1}^K
\rme^{\rmi\bmomega_k\cdot\bmx}
\rme^{-\rmi\bmomega_k\cdot\bmx'}
=: \frac{1}{K}\,\varsigma(\bmx)^\top\varsigma(\bmx')^\ast,
\end{equation}
where
\begin{equation}\label{eq:sigma_inter_represent_gold}
\varsigma(\bmx):=
\big[
\rme^{\rmi\bmomega_1\cdot\bmx},
\rme^{\rmi\bmomega_2\cdot\bmx},
\dots,
\rme^{\rmi\bmomega_K\cdot\bmx}
\big]^\top \in \mathbb C^K.
\end{equation}
Substituting the approximation \eqref{eq:MC_approx_kernel} into the kernel regression model, we obtain
\[
\begin{aligned}
    m_\kappa(\bmx)&=\sum_{i=1}^N \eta_i\, \kappa(\bmx-\bmx_i)\\
    &\simeq \sum_{i=1}^N \eta_i\,\frac{1}{K}\,\varsigma(\bmx)^\top\,{\varsigma(\bmx_i)}^\ast=\varsigma(\bmx)^\top\Big( \sum_{i=1}^N\frac{\eta_i}{K}\,{\varsigma(\bmx_i)}^\ast \Big)=:\varsigma(\bmx)^\top\bm{\beta}\,,
\end{aligned}
\]
where $\bm{\beta}:=\sum_{i=1}^N\frac{\eta_i}{K}\,{\varsigma(\bmx_i)}^\ast\in\mathbb{C}^K$ absorbs the scaling factor. This defines the \textit{random Fourier feature model}:
\begin{equation}\label{eq:RFF_model}
\bar{m}(\bmx)=\bm{\beta}^\top\,\varsigma(\bmx)=\sum_{k=1}^K \beta_k\,\rme^{\rmi{\bmomega}_k\cdot \bmx}\,.
\end{equation}
Equivalently, one may regard \eqref{eq:RFF_model} as a linear model in the explicit random Fourier feature space.

Given sampled frequencies $\{\bmomega_k\}_{k=1}^K$, we estimate the coefficient vector $\bm{\beta}$ from data by solving the regularized least-squares problem
\begin{equation}\label{eq:RFF_lsq}
\min_{\bm{\beta}\in\mathbb C^K}
\left\{
\|\bm{\Phi}\bm{\beta}-\bm{y}\|_2^2
+
\lambda\|\bm{\beta}\|_2^2
\right\},
\end{equation}
where the design matrix $\bm{\Phi}\in\mathbb C^{N\times K}$ is defined by
\[
\Phi_{ik}
=
\rme^{\rmi\bmomega_k\cdot\bmx_i},
\qquad
i=1,\dots,N,\quad k=1,\dots,K\,.
\]
The corresponding normal equations are
\begin{equation}\label{eq:RFF_normal_eq}
(\bm{\Phi}^{\mathrm H}\bm{\Phi}
+
\lambda\bm{I}_K)\bm{\beta}
=
\bm{\Phi}^{\mathrm H}\bm{y}\,,
\end{equation}
where $\bm{\Phi}^{\mathrm H}$ denotes the Hermitian transpose of $\bm{\Phi}$. This reduces the coefficient estimation problem to a regularized least-squares problem in $K$ variables. When $K\ll N$, this provides a computationally tractable approximation to kernel ridge regression.

In our implementation, the initial frequency distribution is refined by an adaptive resampling procedure. Starting from a current set of frequencies $\{\bmomega_k^{(r)}\}_{k=1}^K$, we first fit the RFF coefficients $\bm{\beta}^{(r)}$ by solving \eqref{eq:RFF_lsq}. The fitted amplitudes $|\beta_k^{(r)}|$ are then used to define empirical resampling probabilities
\[
p_k^{(r)}
=
\frac{|\beta_k^{(r)}|}
{\sum_{m=1}^K |\beta_m^{(r)}|},
\qquad k=1,\dots,K\,.
\]
The next set of frequencies is obtained by sampling $K$ frequencies, with replacement, from the current frequency set according to $\{p_k^{(r)}\}_{k=1}^K$. Before each resampling step, the frequencies are perturbed by an adaptive random walk whose covariance is estimated from the previously resampled frequency cloud. This covariance-dependent perturbation allows the algorithm to explore nearby spectral directions while respecting the anisotropy of the learned frequency distribution.

The resampling step adaptively favors spectral components with larger fitted amplitudes, indicating greater contributions to the regression model. Since these amplitudes are estimated from the response data using \eqref{eq:RFF_lsq}, the empirical frequency distribution is tailored to the predictive task. The learned feature map is therefore response-informed, rather than being determined solely by the initial kernel-induced spectral density.

Beyond its role as a global predictor, the trained RFF model also provides a feature map used for latent regime discovery. For each input $\bmx$, we define
\begin{equation}
\label{eq:sx_definition}
s(\bmx)
=
\mathrm{Re}\!\left(\bm{\beta}\odot\varsigma(\bmx)\right)
=
\left[
\mathrm{Re}\!\left(\beta_1\rme^{\rmi\bmomega_1\cdot\bmx}\right),
\dots,
\mathrm{Re}\!\left(\beta_K\rme^{\rmi\bmomega_K\cdot\bmx}\right)
\right]^\top
\in\bbR^K,
\end{equation}
where $\odot$ denotes elementwise multiplication. The entries of $s(\bmx)$ decompose the fitted RFF prediction into spectral contributions.
Since both the amplitudes and the resampled frequencies are determined through the predictive task, $s(\bmx)$ reflects response-relevant structure rather than purely geometric variation in the original covariate space.

Although $K$ is usually chosen to be much smaller than $N$, an accurate random feature approximation may still require a moderately large number of frequencies. Therefore, $s(\bmx)$ can remain high-dimensional and redundant, which motivates the latent compression step below.

\subsubsection{Latent embedding and probabilistic partitioning}
\label{sec:latent_embedding_partitioning}

We compress the response-informed feature vectors before fitting a mixture model. This projection removes redundant spectral directions and produces a latent space in which probabilistic regime discovery is more stable.

For notational clarity, we describe here the projection step with \textit{principal component analysis} (PCA) explicitly. Given the training data $\{(\bmx_i,y_i)\}_{i=1}^N$, we compute the feature map vectors $\{s(\bmx_i)\}_{i=1}^N$ based on \eqref{eq:sx_definition} and form the training feature matrix  $\mathbf{S}\in\bbR^{N\times K}$ by stacking these feature vectors as rows, so that
\[
\mathbf{S}_{ik}
=
\mathrm{Re}\!\left(\beta_k\,\rme^{\rmi\bmomega_k\cdot\bmx_i}\right),
\qquad i=1,\dots,N,\ \ k=1,\dots,K\,.
\]
We then center the feature matrix by subtracting the row-wise empirical mean. The empirical mean feature vector is defined as
\[
\bar{\mathbf{s}}
=
\frac{1}{N}\sum_{i=1}^N s(\bmx_i)
=
\frac{1}{N}\mathbf{S}^\top\mathbf{1}_N
\in\bbR^K,
\]
where $\mathbf{1}_N\in\bbR^N$ denotes the vector of ones. The centered matrix $\bar{\mathbf{S}}\in\bbR^{N\times K}$ is given by
\[
\bar{\mathbf{S}}
=
\mathbf{S}-\mathbf{1}_N\,\bar{\mathbf{s}}^\top.
\]
Next, we compute the singular value decomposition
\[
\bar{\mathbf{S}}=\mathbf{U}\mathbf{\Sigma}\mathbf{V}^\top,
\]
where the columns of $\mathbf{V}\in\bbR^{K\times K}$ record the principal directions of variation in the response-informed feature space, the diagonal entries of $\mathbf{\Sigma}$ are the singular values in descending order, and $\mathbf{U}\in\bbR^{N\times N}$ contains the left singular vectors. Retaining the first $d$ principal directions yields the matrix 
\[
\mathbf{V}_d
=
[\mathbf{v}_1,\dots,\mathbf{v}_d]\in\bbR^{K\times d}\,,
\]
formed by the first $d$ columns $\mathbf{v}_1$, $\dots$, $\mathbf{v}_d$ of the loading matrix $\mathbf{V}$. We construct the $d$-dimensional latent embedding of the training data by
\begin{equation}\label{eq:latent_matrix_Z}
\mathbf{Z}:=\bar{\mathbf{S}}\mathbf{V}_d\in\bbR^{N\times d}\,,
\end{equation}
so that each row of $\mathbf{Z}$ gives the latent coordinates of one training sample.

This projection extracts the dominant modes of variation in the learned spectral representation induced by the fitted RFF model (Hastie et al. \citeyear{ESL_book}). Since the original feature vectors $s(\bmx)$ incorporate response information through both the optimized amplitudes and the adaptive frequency resampling, the resulting latent coordinates summarize the main predictive variation captured by the RFF regression model. Thus, PCA is not used merely as a generic dimensionality-reduction step; rather, it constructs a compact latent space in which predictive regimes can be more stably identified by the subsequent Gaussian mixture model.

For a new input $\bmx$, the same out-of-sample mapping is obtained by first forming $s(\bmx)$, centering it with the empirical mean $\bar{\mathbf{s}}$, and projecting it onto the retained principal directions. We define this latent embedding by
\begin{equation}\label{eq:hx_definition}
h(\bmx)
=
\mathbf{V}_d^\top
\bigl(s(\bmx)-\bar{\mathbf{s}}\bigr)
\in\bbR^d\,.
\end{equation}
For the training data, this definition is consistent with
\eqref{eq:latent_matrix_Z} since $h(\bmx_i)^\top$ equals the
$i$-th row of matrix $\mathbf{Z}$. Across the benchmark regression datasets considered in this paper, a low-dimensional latent space with $d$ between $2$ and $8$ was typically sufficient to support stable and accurate soft clustering; see Table~\ref{tab:hyperparams_summary} in Appendix~\ref{app:baselines} for the selected hyperparameter values across datasets.


Once the spectral map has been compressed into the low-dimensional coordinates $\{\bmz_i\}_{i=1}^N$, where each $\bmz_i\in\bbR^d$ denotes the $i$-th row of the matrix $\mathbf{Z}$ in \eqref{eq:latent_matrix_Z}, we identify latent predictive regimes by fitting a \textit{Gaussian mixture model} (GMM) with $L$ components. This step models the distribution of the PCA-induced latent coordinates rather than partitioning the original covariate space directly, so that regime assignment is based on differences in the learned spectral representation associated with response-relevant variation. Because PCA reduces redundancy and collinearity in the original RFF feature space, the resulting embedding provides a more stable basis for probabilistic regime assignment.

Specifically, we model the latent variable $\bmz$ by the finite mixture density
\begin{equation}
\label{eq:gmm_density}
p(\bmz)
=
\sum_{\ell=1}^L
\pi_\ell\,
\mathcal{N}(\bmz;\bm{\mu}_\ell,\bm{\Sigma}_\ell)\,,
\end{equation}
where $\pi_\ell\geq0$, $\sum_{\ell=1}^L\pi_\ell=1$, and $\mathcal{N}(\bmz;\bm{\mu}_\ell,\bm{\Sigma}_\ell)$ denotes a Gaussian density on $\bbR^d$ with mean vector
$\bm{\mu}_\ell\in\bbR^d$ and covariance matrix
$\bm{\Sigma}_\ell\in\bbR^{d\times d}$. The parameters are estimated by maximum likelihood using the \textit{Expectation-Maximization} (EM) algorithm (Dempster et al. \citeyear{EM_algorithm}; Bishop \citeyear{Bishop_book}).

For a latent point $\bmz$, the posterior responsibility of component $\ell$ is
\begin{equation}
\label{eq:gmm_responsibility}
\gamma_\ell(\bmz)
=
\frac{
\pi_\ell\,
\mathcal{N}(\bmz;\bm{\mu}_\ell,\bm{\Sigma}_\ell)
}{
\sum_{\ell'=1}^L
\pi_{\ell'}\,
\mathcal{N}(\bmz;\bm{\mu}_{\ell'},\bm{\Sigma}_{\ell'})
}\,.
\end{equation}
Evaluating this responsibility at $h(\bmx)$ induces the input-dependent regime weight
\begin{equation}
\label{eq:regime_weight}
w_\ell(\bmx)
=
\gamma_\ell\big(h(\bmx)\big),
\qquad
\ell=1,\dots,L\,.
\end{equation}
These weights are nonnegative and sum to one.

The GMM therefore converts the response-informed latent embedding into a soft partition of the input space. The same posterior responsibilities also provide observation-level weights for fitting local GAMs. When regimes are well separated, they may instead be converted into hard assignments by selecting the component with maximal responsibility. The next subsection formulates both choices within a single weighted additive-model objective.

\subsection{Cluster-wise generalized additive modeling}
\label{sec:clusterwise_gam}

After identifying latent predictive regimes, we model the regression structure within each regime using generalized additive models (Hastie and Tibshirani \citeyear{GAM_book}; Hodges \citeyear{Hodges_book};
Wood \citeyear{Wood_book}). The purpose of this stage is to translate the
component structure detected in the response-informed latent embedding into
interpretable local regression models defined on the original covariates.

The posterior responsibilities obtained from the Gaussian mixture model provide a natural mechanism for connecting the latent partition with the local additive models. Recalling from \eqref{eq:regime_weight} the input-dependent regime weight, for training the local GAMs, we associate each observation $(\bmx_i,y_i)$ and each component $\ell$ with a nonnegative fitting weight $a_{i\ell}$, which measures how strongly that observation contributes to the estimation of the $\ell$-th local model.

In the soft formulation, these fitting weights are taken directly as the posterior responsibilities:
\[
a_{i\ell}^{\mathrm{soft}}
=
\gamma_\ell(\bmz_i),
\quad
\textup{with } \ \;\bmz_i=h(\bmx_i)\,.
\]
This allows an observation to contribute to multiple local models according to its posterior membership in the latent space. Such a formulation is useful when regime boundaries are gradual or when observations near the boundary carry information about more than one component.

A hard-clustering formulation is obtained by assigning each observation to the
component with the largest posterior responsibility. Specifically, we can define
\[
c_i
=
\argmax_{1\le r\le L}\,\gamma_r(\bmz_i),
\]
and set
\[
a_{i\ell}^{\mathrm{hard}}
=
\mathbf{1}\{c_i=\ell\},
\]
where $\mathbf{1}\{\cdot\}$ denotes the indicator function. This formulation may be preferable when the latent regimes are well separated, since it prevents observations from one regime from influencing the estimation of neighboring local models.

In what follows, we write $a_{i\ell}$ generically for the fitting weights, allowing either the soft choice $a_{i\ell}=a_{i\ell}^{\mathrm{soft}}$ or the hard choice $a_{i\ell}=a_{i\ell}^{\mathrm{hard}}$. Both choices are derived from the same fitted Gaussian mixture model and therefore preserve the response-informed regime structure.

For each component $\ell$, we fit a generalized additive model of the form
\[
\tilde{f}^{(\ell)}(\bmx)
=
\alpha^{(\ell)}+\sum_{j=1}^p g_j^{(\ell)}(x_j),
\]
where $\alpha^{(\ell)}$ is an intercept and $g_j^{(\ell)}$ is a smooth univariate function describing the effect of covariate $x_j$ within the $\ell$-th regime. Each component function is represented by a
B-spline expansion (de Boor \citeyear{Spline_book}),
\[
g_j^{(\ell)}(x_j)
=
\sum_{q=1}^{Q_j}
\theta_{j,q}^{(\ell)}\,\phi_{j,q}(x_j),
\]
where $\{\phi_{j,q}\}_{q=1}^{Q_j}$ are spline basis functions constructed using
quantile-based interior knots, and $\theta_{j,q}^{(\ell)}$ are the corresponding
coefficients.

The parameters of the $\ell$-th local GAM are estimated by solving the weighted
penalized least-squares problem
\begin{equation}\label{eq:weighted_local_GAMs}
\min_{\alpha^{(\ell)},\,\{g_j^{(\ell)}\}_{j=1}^p}
\Big\{
\sum_{i=1}^{N}
a_{i\ell}
\big(
y_i-\alpha^{(\ell)}-\sum_{j=1}^p g_j^{(\ell)}(x_{ij})
\big)^2
+
\lambda_{\mathrm{GAM}}
\sum_{j=1}^p
\int
\big(g_j^{(\ell)\,\prime\prime}(t)\big)^2
\,\mathrm{d}t
\Big\}\,.
\end{equation}
where $x_{ij}$ denotes the $j$-th covariate of the observation $\bmx_i$, and
$\lambda_{\mathrm{GAM}}>0$ controls the smoothness of the fitted additive
components. In practice, after substituting the B-spline expansion, the roughness
penalty is implemented through the quadratic approximation
\[
\int
\big(g_j^{(\ell)\,\prime\prime}(t)\big)^2
\,\mathrm{d}t
\approx
(\bm{\theta}_j^{(\ell)})^\top
\bm{\Omega}_j
\bm{\theta}_j^{(\ell)}\,,
\]
where
\[
\bm{\theta}_j^{(\ell)}
=
(\theta_{j,1}^{(\ell)},\dots,\theta_{j,Q_j}^{(\ell)})^\top
\]
is the coefficient vector for the $j$-th smooth function in component $\ell$, and
$\bm{\Omega}_j$ is the penalty matrix associated with second-order differences of the spline coefficients.

The weighted formulation \eqref{eq:weighted_local_GAMs} is important for connecting the latent partition to the local additive models. When the weights are chosen as posterior responsibilities, the criterion yields a soft, cluster-wise GAM-fitting procedure: observations near regime boundaries can influence multiple local models, with the strength determined by their posterior membership probabilities. If the weights are instead binary indicators, the same objective reduces to a hard-clustering formulation in which each observation is assigned to only one local model. Thus, the weighted criterion provides a unified formulation encompassing both hard and soft regime-specific GAM fitting.

This local additive structure is central to the interpretability of the proposed
framework. The latent-space mixture model identifies regimes in which the predictive mechanism may differ, while the GAM fitted within each regime decomposes that mechanism into covariate-specific smooth effects. Consequently, each function $g_j^{(\ell)}$ can be interpreted as the contribution of covariate $x_j$ to the response within regime $\ell$. These component-wise effects can be visualized through the fitted smooth functions, enabling interpretation at both the regime level and the covariate level. The local models are subsequently combined using the same input-dependent weights $w_\ell(\bmx)$, as described in the overall prediction pipeline in the following section.

\subsection{Integrated prediction pipeline}
\label{sec:prediction_pipeline}

The preceding subsections define the response-informed representation, the latent regime weights, and the cluster-wise additive models. We now combine these components into the final predictor.

For a new input $\bmx$, the trained RFF model produces spectral feature map $s(\bmx)$, the PCA map gives latent embedding $h(\bmx)$, and the GMM assigns regime weights $w_\ell(\bmx)=\gamma_\ell\big(h(\bmx)\big)$. The final mixture-of-GAMs predictor is
\begin{equation}
\label{eq:final_mixture_gam}
\tilde m(\bmx)
=
\sum_{\ell=1}^L
\gamma_\ell\big(h(\bmx)\big)\,
\tilde f^{(\ell)}(\bmx).
\end{equation}
Thus, prediction is obtained by aggregating interpretable local additive models using regime weights derived from the response-informed latent representation.

\begin{figure}[htbp]
\centering
\resizebox{\textwidth}{!}{
\begin{tikzpicture}[
  font=\small,
  node distance=12mm and 12mm,
  box/.style={rounded corners, draw, very thick, inner sep=6pt, align=left, fill=white, minimum width=50mm},
  line/.style={-Latex, very thick},
  title/.style={font=\bfseries},
  note/.style={font=\itshape, inner sep=0pt},
  var/.style={font=\ttfamily\footnotesize}
]

\node[box, fill=blue!4] (rff) {\textbf{Stage 1: Train RFF model}\\
\textcolor{blue!80!black}{$\displaystyle \bar m(\bmx)=\bm{\beta}^{\top}\varsigma(\bmx)$}\\[1mm]
\textbf{Data:} $\{(\bmx_i,y_i)\}_{i=1}^N$\\
\textbf{Parameters:} $\{\bmomega_k\}_{k=1}^K$\\
\textbf{Learns:} $\bm{\beta}\in\mathbb{C}^K$\\
\textbf{Builds:} $\mathbf{S}\in\mathbb{R}^{N\times K}$,\; $\mathbf{S}_{ik}=\mathrm{Re}(\beta_k \,\rme^{\mathrm{i}\bmomega_k\cdot \bmx_i})$};

\node[box, fill=orange!5, right=25mm of rff] (pca) {\textbf{Stage 2: PCA \& GMM clustering}\\
Center $\mathbf{S}\to \bar{\mathbf{S}}$, SVD: $\bar{\mathbf{S}}=\mathbf{U}\mathbf{\Sigma} \mathbf{V}^\top$\\
\textbf{Keeps:} $\mathbf{V}_d$,\; $\mathbf{Z}=\bar{\mathbf{S}} \mathbf{V}_d\in\mathbb{R}^{N\times d}$\\[0.5mm]
{Fit GMM} on $\mathbf{Z}$: $p(\bmz)=\sum_{\ell=1}^L \pi_\ell\,\mathcal{N}(\bmz;\mu_\ell,\Sigma_\ell)$\\
\textbf{Posteriors:} $\displaystyle \gamma_\ell(h(\bmx))=\frac{\pi_\ell\,\mathcal{N}(h(\bmx);\mu_\ell,\Sigma_\ell)}
{\sum_{\ell'} \pi_{\ell'}\,\mathcal{N}(h(\bmx);\mu_{\ell'},\Sigma_{\ell'})}$};

\node[box, fill=purple!6, below=22mm of rff] (gams) {\textbf{Stage 3: Fit Local GAMs}\\
For $\ell=1,\dots,L$:\\
$\displaystyle \tilde f^{(\ell)}(\bmx)=\alpha^{(\ell)}+\sum_{j=1}^p g^{(\ell)}_j(x_j)$\\
$g^{(\ell)}_j(x_j)=\sum_{q=1}^{Q_j}\theta^{(\ell)}_{j,q}\,\phi_{j,q}(x_j)$
\vspace{1.5mm}
\\ \textbf{Trains on:} cluster-assigned subsets};

\node[box, fill=green!4, right=25mm of gams] (mix) {\textbf{Stage 4: Mixture Prediction}\\
For new $\bmx$: $s(\bmx)=\mathrm{Re}(\bm{\beta}\odot \varsigma(\bmx))$\\
$h(\bmx)=\mathbf{V}_d^\top\big(s(\bmx)-\bar{\mathbf{s}}\big)$\\[0.5mm]
\textbf{Output:} \textcolor{RedOrange}{$\displaystyle \tilde m(\bmx)=\sum_{\ell=1}^L \gamma_\ell\big(h(\bmx)\big)\,\tilde f^{(\ell)}(\bmx)$}};

\draw[line] (rff) -- (pca); 
\draw[line] (pca.east) -- ([xshift=5mm]pca.east) |- ([yshift=5mm]gams.north) -| ([xshift=-5mm]gams.west) -- (gams.west);
\draw[line] (gams) -- (mix); 

\node[below=3mm of rff, var]  (v1) {$\varsigma(\bmx)=[\rme^{\mathrm{i}\bmomega_1\cdot \bmx},\dots,\rme^{\mathrm{i}\bmomega_K\cdot \bmx}]^{\top}$,\quad $\bar{\mathbf{s}}=\tfrac{1}{N}\sum_{i=1}^N \mathbf{S}_{i,:}$};
\node[below=3mm of pca, var]  (v2) {$\mathbf{V}_d\in\mathbb{R}^{K\times d}$,\  $\mathbf{Z}\in\mathbb{R}^{N\times d}$,\ $h(\bmx): =\mathbf{V}_d^\top\big(\mathrm{Re}\big(\bm{\beta}\odot\varsigma(\bmx)\big)-\bar{\mathbf{s}}\big)$};
\node[below=3mm of gams, var]  (v3) {$[x_1,\dots,x_p]=\bmx\in\bbR^p$};
\node[below=3mm of mix, var]  (v4) {$\bm{\beta}\in\mathbb{C}^K$,\; $\odot$: Hadamard product};
\end{tikzpicture}
}
\caption{The overall pipeline of the RFF-informed mixture-of-GAMs. Stage 1 learns RFF model coefficients
and builds a random Fourier feature space representation. Stage 2 reduces to a lower-dimensional embedding via PCA and fits a GMM to obtain posterior responsibilities.
Stage 3 trains local GAMs for each cluster of the training dataset. Stage 4 forms the final prediction.}
\label{fig:workflow_mixture_gam_2x2}
\end{figure}
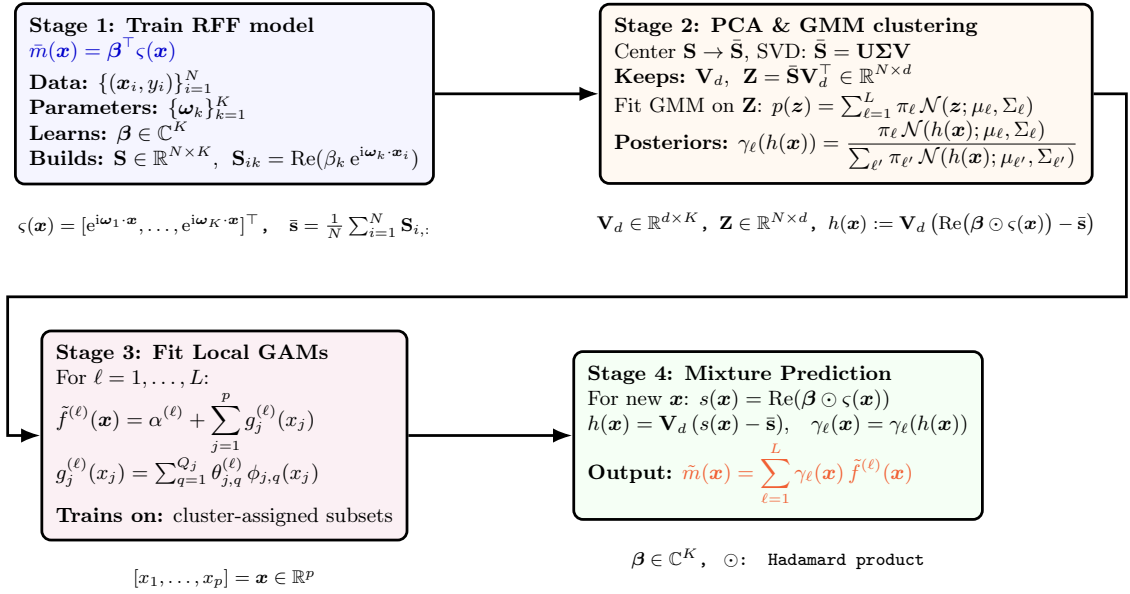

\begin{figure}[htbp]
    \centering
    \includegraphics[width=1.05\linewidth]{ 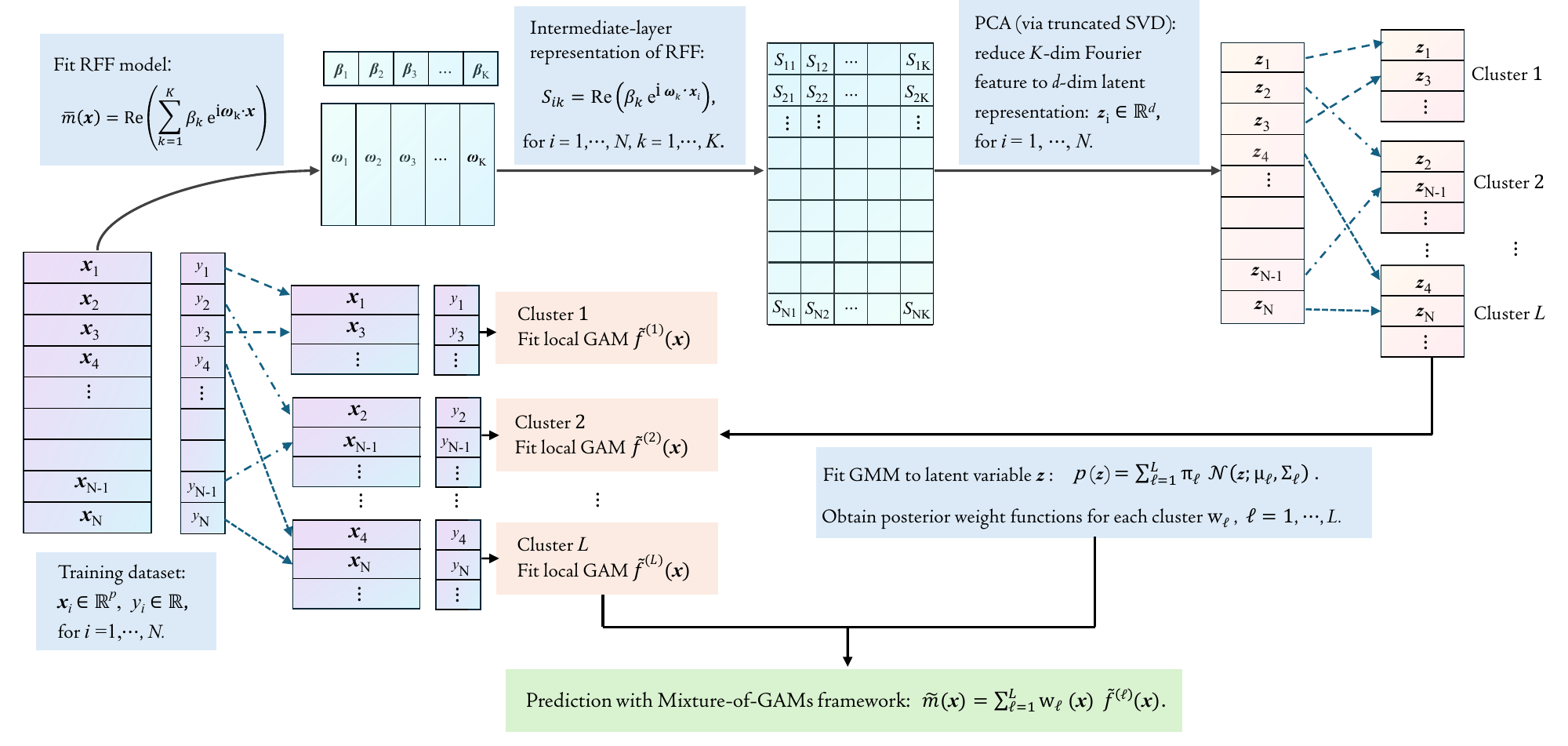}
    \caption{Diagram with graphical representations of the workflow of the mixture-of-GAMs method informed with random Fourier features.}
    \label{fig:workflow_plot}
\end{figure}

The complete training and prediction procedure is summarized in Algorithm~\ref{alg:mixture_GAM_workflow}. Figure~\ref{fig:workflow_mixture_gam_2x2} illustrates the overall modeling pipeline, and Figure~\ref{fig:workflow_plot} gives a detailed schematic of the matrix construction, latent projection, and mixture-based prediction steps.

\begin{algorithm}[h!]
\caption{Training Pipeline for RFF-informed Mixture-of-GAMs}
\label{alg:mixture_GAM_workflow}
\begin{algorithmic}
\STATE \textbf{Input:} Training data $\{(\bmx_i, y_i)\}_{i=1}^N$, number of Fourier features $K$, reduced dimension $d$, number of clusters $L$
\vspace{0.3em}
\STATE \textbf{Output:} Trained local GAM models $\{\tilde{f}^{(\ell)}(\bmx)\}_{\ell=1}^L$ and posterior weight functions $\{\gamma_\ell\big(h(\bmx)\big)\}_{\ell=1}^L$
\vspace{0.4em}
\STATE \textbf{Stage 1: Training of Random Fourier Feature model}
\vspace{-0.35em}
\begin{itemize}\setlength{\itemsep}{0.3pt}
    \item Sample frequency parameters $\{\bmomega_k\}_{k=1}^K \sim \rho(\bmomega)$
    \item Construct feature mapping $\varsigma(\bmx) := [\rme^{\rmi \bmomega_1 \cdot \bmx}, \dots, \rme^{\rmi \bmomega_K \cdot \bmx}]^\top$
    \item Train the RFF regression model $\bar{m}(\bmx) = \bm{\beta}^\top \varsigma(\bmx)$ 
    \item Define intermediate feature vector $s(\bmx) := \mathrm{Re}(\bm{\beta} \odot \varsigma(\bmx))$
    \item Build intermediate feature matrix $\mathbf{S} \in \mathbb{R}^{N \times K}$ by stacking rows $s(\bmx_i)^\top$ for $i=1,\dots,N$
\end{itemize}
\vspace{-0.4em}
\STATE \textbf{Stage 2: Dimensionality Reduction and GMM-based Clustering}
\vspace{-0.4em}
\begin{itemize}\setlength{\itemsep}{0.3pt}
    \item Obtain centered feature matrix $\bar{\mathbf{S}}$ by subtracting from each row of $\mathbf{S}$ the empirical mean $\bar{\mathbf{s}}=\frac{1}{N}\sum_{i=1}^N s(\bmx_i)$.
    \item Apply PCA via SVD: $\bar{\mathbf{S}} = \mathbf{U}\bm{\Sigma}\mathbf{V}^\top$, retain first $d$ principal directions $\mathbf{V}_d$.
    \item Compute reduced representation: $\mathbf{Z} = \bar{\mathbf{S}}\mathbf{V}_d \in \mathbb{R}^{N \times d}$, with rows $\bmz_i=h(\bmx_i)^\top$.
    \item Fit a GMM on $\mathbf{Z}$ to obtain mixture proportions $\{\pi_\ell\}_{\ell=1}^L$, means $\{\bm{\mu}_\ell\}_{\ell=1}^L$, and covariances $\{\bm{\Sigma}_\ell\}_{\ell=1}^L$.
    \item For each mixture component $\ell$, derive posterior responsibilities $\gamma_\ell(h(\bmx))=\frac{\pi_\ell\,\mathcal{N}(h(\bmx);\bm{\mu}_\ell,\bm{\Sigma}_\ell)}{\sum_{\ell'=1}^L \pi_{\ell'}\,\mathcal{N}(h(\bmx);\bm{\mu}_{\ell'},\bm{\Sigma}_{\ell'})}$ , where $h(\bmx)=\mathbf{V}_d^\top(s(\bmx)-\bar{\mathbf{s}})$.
    \item Define fitting weights $a_{i\ell}$ for local GAM training. In the soft formulation, set $a_{i\ell}=\gamma_\ell(\bmz_i)$; in the hard formulation, set $c_i=\arg\max_{1\le r\le L}\gamma_r(\bmz_i)$ and $a_{i\ell}=\mathbf{1}\{c_i=\ell\}$.
\end{itemize}

\vspace{-0.4em}
\STATE \textbf{Stage 3: Training of Local GAMs}
\vspace{0.05em}
\\For each $\ell = 1,\dots,L$:
\vspace{-0.45em}
\begin{itemize}\setlength{\itemsep}{0.3pt}
    \item Using the fitting weights $\{a_{i\ell}\}_{i=1}^N$, fit a weighted local GAM :
    $\tilde{f}^{(\ell)}(\bmx)=\alpha^{(\ell)}+\sum_{j=1}^p g_j^{(\ell)}(x_j)$.
    \item Represent each additive component $g_j^{(\ell)}$ using a B-spline basis with quantile-based knots, and estimate the local model by weighted penalized GAM fitting.
\end{itemize}
\vspace{-0.3em}
\STATE \textbf{Stage 4: Mixture-of-GAMs Prediction}
\vspace{0.05em}
\\For a new input $\bmx$:
\vspace{-0.45em}
    \begin{itemize}\setlength{\itemsep}{0.3pt}
        \item Compute $s(\bmx) = \mathrm{Re}(\bm{\beta} \odot \varsigma(\bmx))$
        \item Compute reduced embedding $h(\bmx) = \mathbf{V}_d^\top\,(s(\bmx) - \bar{\mathbf{s}})$
        \item Evaluate responsibilities $\{\gamma_\ell\big(h(\bmx)\big)\}_{\ell=1}^L$
        \item Predict: 
        \[
        \tilde{m}(\bmx) = \sum_{\ell=1}^L \gamma_\ell\big(h(\bmx)\big)\, \tilde{f}^{(\ell)}(\bmx)
        \]
    \end{itemize}

\end{algorithmic}
\end{algorithm}

\section{Numerical Experiment}\label{Sec_4}
To numerically validate the proposed mixture-of-GAMs framework, we apply Algorithm~\ref{alg:mixture_GAM_workflow} to six benchmark regression datasets drawn from different application domains, each exhibiting varying degrees of nonlinearity and heterogeneity. Across these datasets, we compare the proposed approach with a range of baseline models spanning both classical statistical methods and modern machine learning techniques. The baselines include multilayer perceptron (MLP, Rumelhart et al. \citeyear{back_propagation_paper}), Random Forest (Breiman \citeyear{Random_forest}), XGBoost (Chen and Guestrin \citeyear{XGBoost}), random Fourier feature (RFF) regression with resampling-based frequency optimization (Huang et al. \citeyear{RFF_resampling}), global generalized additive model (GAM, Hastie and Tibshirani \citeyear{GAM_book}), least absolute shrinkage and selection operator (LASSO, Tibshirani \citeyear{Lasso}), multivariate adaptive regression splines (MARS, Friedman \citeyear{MARS}), Explainable Boosting Machine (Caruana et al. \citeyear{Interpretable_Boost}; Nori et al. \citeyear{Explainable_Boost}), and mixture-of-linear-models (MLM, Seo et al. \citeyear{MLM}). These methods are grouped into three categories: highly expressive but less directly interpretable models (MLP, Random Forest, XGBoost, and RFF), standard interpretable regression approaches (LASSO, MARS, global GAM, and Explainable Boosting Machine,), and locally adaptive mixture models based on either linear components (MLM) or generalized additive components (mixture-of-GAMs).

\subsection{Experimental setup and data description}\label{subsec_numerical_setup}

We evaluate the proposed framework on six benchmark regression datasets: California Housing (Pace and Barry \citeyear{pace1997sparse}), Airfoil Self-Noise (Brooks et al. \citeyear{Airfoil_dataset}), Bike Sharing (Fanaee-T and Gama \citeyear{bikesharing_dataset}), Kin40k (Rasmussen and Williams \citeyear{GPML_book}), Elevators (Dua and Graff \citeyear{uci_repository}), and Protein (Moult et al. \citeyear{CASP}). Their key characteristics are summarized in Table~\ref{tab:dataset_summary}. These datasets represent a diverse range of regression tasks, varying in sample size, dimensionality, and structural complexity, while encompassing both smooth and highly nonlinear regimes.

All datasets are split into training and testing sets using the provided partitions or fixed random splits. Input features are standardized based on training statistics. Model performance is evaluated using the root mean squared error (RMSE), with uncertainty quantified via residual bootstrap confidence intervals based on 1000 resamples.

The proposed mixture-of-GAMs framework involves two key hyperparameters: the dimension $d$ of the PCA-induced latent embedding and the number of mixture components $L$. 
For each dataset, we select the values of $d$ and $L$ by grid search based on predictive performance on the validation data. 
Detailed hyperparameter settings for the proposed method, including random Fourier feature configurations, clustering parameters, spline specifications, and additional visualization of the corresponding performance landscape over $(d,L)$ pairs are provided in Appendix~\ref{app:implementation}.

\begin{table}[htbp]
\centering
\caption{Summary of benchmark datasets used in the experiments.}
\label{tab:dataset_summary}
\begin{tabular}{lcccc}
\toprule
\textbf{Dataset} & \textbf{Train size} & \textbf{Test size} & \textbf{\# Covariates} & \textbf{Response variable} \\
\midrule
California Housing & 16512 & 4128 & 8 & Median house value ($10^5$ USD) \\
Airfoil Self-Noise & 1202 & 301 & 5 & Sound pressure level (dB) \\
Bike Sharing & 13903 & 3476 & 13 & Hourly rental count \\
Kin40k & 36000 & 4000 & 8 & Forward kinematics output \\
Elevators & 14940 & 1659 & 18 & Control signal \\
Protein & 41157 & 4573 & 9 & Root mean square deviation \\
\bottomrule
\end{tabular}
\end{table}

\subsection{Prediction performance}
\subsubsection{Overall predictive accuracy across datasets}
\label{sec:predictive_performance_overall}

The primary objective of this subsection is to assess how the proposed mixture-of-GAMs framework navigates the trade-off between predictive accuracy and interpretability across a diverse collection of regression settings. Figure~\ref{fig:rmse_bar_all_datasets} summarizes the training and test RMSE values for nine representative methods across six benchmark datasets, spanning highly expressive black-box predictors, globally interpretable regression models, and the proposed locally adaptive framework. 

\begin{figure}[htbp]
\centering
\begin{tikzpicture}

\begin{groupplot}[
    group style={
        group size=1 by 6,
        vertical sep=0.55cm
    },
    ybar,
    /pgf/bar width=8.5pt,
    width=\textwidth,
    height=3.7cm,
    symbolic x coords={MLP,Random Forest,XGBoost,RFF,LASSO,MARS,Global GAM,Expl. Boost. Mach.,Mixture of GAMs},
    xtick=data,
    enlarge x limits=0.06,
    ylabel={RMSE},
    xticklabel style={rotate=45, anchor=east, font=\scriptsize},
    ymajorgrids=true,
    grid style={dashed,gray!30},
]

\nextgroupplot[
    ymin=0.15, ymax=0.80,
    xticklabels={,,,,,,,,},
]
\node[
    anchor=north west,
    draw=black,
    rounded corners=1pt,
    fill=white,
    font=\small,
    inner sep=2pt
] at (rel axis cs:0.02,0.95) {California Housing};

\addplot[
    fill=red!30,
    draw=red!60
] coordinates {
    (MLP,0.503)
    (Random Forest,0.181)
    (XGBoost,0.236)
    (RFF,0.396)
    (LASSO,0.726)
    (MARS,0.629)
    (Global GAM,0.548)
    (Expl. Boost. Mach.,0.453)
    (Mixture of GAMs,0.442)
};

\addplot+[
    fill=blue!30,
    draw=blue!60,
    error bars/.cd,
        y dir=both,
        y explicit,
        error bar style={blue!70},
] coordinates {
    (MLP,0.517) +- (0,0.035)
    (Random Forest,0.502) +- (0,0.023)
    (XGBoost,0.437) +- (0,0.021)
    (RFF,0.439) +- (0,0.021)
    (LASSO,0.723) +- (0,0.023)
    (MARS,0.640) +- (0,0.022)
    (Global GAM,0.567) +- (0,0.023)
    (Expl. Boost. Mach.,0.489) +- (0,0.023)
    (Mixture of GAMs,0.489) +- (0,0.021)
};


\nextgroupplot[
    ymin=0.0, ymax=5.4,
    xticklabels={,,,,,,,,},
]
\node[
    anchor=north west,
    draw=black,
    rounded corners=1pt,
    fill=white,
    font=\small,
    inner sep=2pt
] at (rel axis cs:0.02,0.95) {Airfoil Self-Noise};

\addplot[
    fill=red!30,
    draw=red!60
] coordinates {
    (MLP,3.76)
    (Random Forest,0.68)
    (XGBoost,0.36)
    (RFF,0.67)
    (LASSO,4.79)
    (MARS,4.95)
    (Global GAM,4.45)
    (Expl. Boost. Mach.,1.38)
    (Mixture of GAMs,1.84)
};

\addplot+[
    fill=blue!30,
    draw=blue!60,
    error bars/.cd,
        y dir=both,
        y explicit,
        error bar style={blue!70},
] coordinates {
    (MLP,3.80) +- (0,0.04)
    (Random Forest,1.60) +- (0,0.03)
    (XGBoost,1.43) +- (0,0.03)
    (RFF,1.08) +- (0,0.02)
    (LASSO,4.82) +- (0,0.09)
    (MARS,4.98) +- (0,0.08)
    (Global GAM,4.51) +- (0,0.08)
    (Expl. Boost. Mach.,2.21) +- (0,0.04)
    (Mixture of GAMs,2.02) +- (0,0.04)
};

\nextgroupplot[
    ymin=0, ymax=155,
    xticklabels={,,,,,,,,},
]

\node[
    anchor=north west,
    draw=black,
    rounded corners=1pt,
    fill=white,
    font=\small,
    inner sep=2pt
] at (rel axis cs:0.02,0.95) {Bike Sharing};

\addplot[
    fill=red!30,
    draw=red!60
] coordinates {
    (MLP,41.3)
    (Random Forest,58.3)
    (XGBoost,24.8)
    (RFF,87.5)
    (LASSO,141.1)
    (MARS,120.9)
    (Global GAM,85.9)
    (Expl. Boost. Mach.,56.4)
    (Mixture of GAMs,48.6)
};

\addplot+[
    fill=blue!30,
    draw=blue!60,
    error bars/.cd,
        y dir=both,
        y explicit,
        error bar style={blue!70},
] coordinates {
    (MLP,47.4) +- (0,3.0)
    (Random Forest,72.6) +- (0,3.3)
    (XGBoost,37.7) +- (0,2.8)
    (RFF,92.3) +- (0,3.5)
    (LASSO,140.3) +- (0,5.0)
    (MARS,122.5) +- (0,4.5)
    (Global GAM,88.8) +- (0,5.6)
    (Expl. Boost. Mach.,57.7) +- (0,2.5)
    (Mixture of GAMs,52.5) +- (0,2.4)
};

\nextgroupplot[
    ymin=0.0, ymax=1.08,
    xticklabels={,,,,,,,,},
]

\node[
    anchor=north west,
    draw=black,
    rounded corners=1pt,
    fill=white,
    font=\small,
    inner sep=2pt
] at (rel axis cs:0.02,0.95) {Kin40k};

\addplot[
    fill=red!30,
    draw=red!60
] coordinates {
    (MLP,0.059)
    (Random Forest,0.388)
    (XGBoost,0.206)
    (RFF,0.066)
    (LASSO,1.000)
    (MARS,0.972)
    (Global GAM,0.985)
    (Expl. Boost. Mach.,0.811)
    (Mixture of GAMs,0.305)
};

\addplot+[
    fill=blue!30,
    draw=blue!60,
    error bars/.cd,
        y dir=both,
        y explicit,
        error bar style={blue!70},
] coordinates {
    (MLP,0.071) +- (0,0.003)
    (Random Forest,0.494) +- (0,0.013)
    (XGBoost,0.358) +- (0,0.011)
    (RFF,0.071) +- (0,0.004)
    (LASSO,0.971) +- (0,0.021)
    (MARS,0.953) +- (0,0.022)
    (Global GAM,0.961) +- (0,0.021)
    (Expl. Boost. Mach.,0.810) +- (0,0.021)
    (Mixture of GAMs,0.339) +- (0,0.012)
};

\nextgroupplot[
    ymin=0.0, ymax=0.26,
    xticklabels={,,,,,,,,},
]

\node[
    anchor=north west,
    draw=black,
    rounded corners=1pt,
    fill=white,
    font=\small,
    inner sep=2pt
] at (rel axis cs:0.02,0.95) {Elevators};

\addplot[
    fill=red!30,
    draw=red!60
] coordinates {
    (MLP,0.083)
    (Random Forest,0.097)
    (XGBoost,0.065)
    (RFF,0.081)
    (LASSO,0.123)
    (MARS,0.105)
    (Global GAM,0.227)
    (Expl. Boost. Mach.,0.091)
    (Mixture of GAMs,0.082)
};

\addplot+[
    fill=blue!30,
    draw=blue!60,
    error bars/.cd,
        y dir=both,
        y explicit,
        error bar style={blue!70},
] coordinates {
    (MLP,0.093) +- (0,0.004)
    (Random Forest,0.125) +- (0,0.007)
    (XGBoost,0.095) +- (0,0.005)
    (RFF,0.088) +- (0,0.004)
    (LASSO,0.126) +- (0,0.006)
    (MARS,0.107) +- (0,0.004)
    (Global GAM,0.231) +- (0,0.012)
    (Expl. Boost. Mach.,0.096) +- (0,0.004)
    (Mixture of GAMs,0.089) +- (0,0.004)
};

\nextgroupplot[
    ymin=0.0, ymax=0.72,
]

\node[
    anchor=north west,
    draw=black,
    rounded corners=1pt,
    fill=white,
    font=\small,
    inner sep=2pt
] at (rel axis cs:0.02,0.95) {Protein};

\addplot[
    fill=red!30,
    draw=red!60
] coordinates {
    (MLP,0.278)
    (Random Forest,0.345)
    (XGBoost,0.302)
    (RFF,0.311)
    (LASSO,0.654)
    (MARS,0.628)
    (Global GAM,0.642)
    (Expl. Boost. Mach.,0.471)
    (Mixture of GAMs,0.476)
};

\addplot+[
    fill=blue!30,
    draw=blue!60,
    error bars/.cd,
        y dir=both,
        y explicit,
        error bar style={blue!70},
] coordinates {
    (MLP,0.427) +- (0,0.016)
    (Random Forest,0.434) +- (0,0.013)
    (XGBoost,0.419) +- (0,0.014)
    (RFF,0.390) +- (0,0.013)
    (LASSO,0.655) +- (0,0.011)
    (MARS,0.630) +- (0,0.013)
    (Global GAM,0.641) +- (0,0.012)
    (Expl. Boost. Mach.,0.509) +- (0,0.013)
    (Mixture of GAMs,0.498) +- (0,0.014)
};

\end{groupplot}

\node[anchor=north east] at ($(current bounding box.north east)+(0.1cm,0.9cm)$) {
\begin{tikzpicture}
\begin{axis}[
    hide axis,
    xmin=0, xmax=1,
    ymin=0, ymax=1,
    legend style={draw=none, fill=none, font=\small},
    legend image code/.code={
        \draw[#1, fill] (0cm,-0.08cm) rectangle (0.12cm,0.18cm);
    }
]
\addlegendimage{fill=red!30,draw=red!60}
\addlegendentry{Training RMSE}

\addlegendimage{fill=blue!30,draw=blue!60}
\addlegendentry{Test RMSE (95\% CI)}
\end{axis}

\end{tikzpicture}
};

\end{tikzpicture}
\caption{Comparison of training and test RMSE across six benchmark datasets for nine model classes. Test RMSE bars include 95\% confidence intervals obtained via bootstrap resampling.}
\label{fig:rmse_bar_all_datasets}
\end{figure}

\paragraph{Flexible predictive reference level.}
Several consistent patterns emerge among the highly expressive baselines, including MLP, Random Forest, XGBoost, and the resampling-based RFF model. Across most datasets, the RFF model is among the strongest performers and in several cases is competitive with or superior to the neural-network and tree-based alternatives. This behavior is particularly evident in the California Housing, Airfoil Self-Noise, Kin40k, Elevators, and Protein datasets, whereas the Bike Sharing benchmark uses a deliberately restricted temporal RFF specification to support the interpretability analysis discussed in Section~\ref{subsection_regime}. These results indicate that the learned spectral representation provides an effective approximation of the underlying regression structure. Accordingly, the best-performing flexible models in Figure~\ref{fig:rmse_bar_all_datasets} provide a useful reference level of predictive accuracy when direct structural interpretability is not the primary modeling constraint.

\paragraph{Recovering performance through local adaptivity.}
Against this reference point, the central empirical question is whether predictive performance can be recovered within a substantially more interpretable model class by allowing localized structure rather than relying on a single global predictor. Standard globally interpretable baselines, including LASSO, MARS, and the global GAM, generally exhibit noticeably larger test errors on datasets characterized by stronger nonlinearity, regime variation, or non-additive interactions.

The proposed mixture-of-GAMs framework consistently achieves an improvement on these global baselines across all six datasets, supporting the value of combining response-informed regime discovery with interpretable local additive modeling. The gains are particularly pronounced on datasets where heterogeneous predictive mechanisms are crucial. On the Bike Sharing dataset, the proposed method achieves a substantial reduction in RMSE relative to the global GAM and the Explainable Boosting Machine, indicating that local specialization is beneficial when demand patterns vary across temporal regimes such as commuting, leisure, and seasonal usage periods. On Kin40k, which is governed by nonlinear trigonometric transformations and strong cross-variable interactions, the mixture of GAMs also clearly outperforms global additive baselines. These observations suggest that the RFF-guided latent partitioning enables a complex response surface to be approximated by a soft combination of simpler local additive components.

\paragraph{Quantifying the predictive cost of interpretability.}
While the proposed framework substantially narrows the gap between global interpretable models and highly flexible predictors, Figure~\ref{fig:rmse_bar_all_datasets} also reveals the remaining predictive cost of enforcing a locally additive structure. This gap is most evident in datasets such as Airfoil Self-Noise, Kin40k, and Protein, where the strongest black-box or spectral predictors, especially the RFF model, retain a noticeable performance advantage.

This remaining discrepancy is scientifically informative. It indicates that even after introducing latent regimes, a model based on univariate additive components may still face limitations when strongly coupled, high-dimensional physical relationships govern the underlying system. In such settings, the primary advantage of the proposed framework is therefore not absolute predictive capacity, but a more favorable balance between transparency and predictive strength  than that offered by conventional global interpretable models.

\paragraph{Generalization and structural stability.}
A further observation concerns the relationship between training and test accuracy. Several flexible baselines, particularly Random Forest and XGBoost, often exhibit larger gaps between training and test RMSE, reflecting their greater tendency to fit the training data aggressively. By contrast, the mixture of GAMs typically maintains a more stable profile while still achieving competitive test accuracy. The combination of response-informed clustering and locally regularized additive modeling provides an effective form of structural control, allowing the model to capture essential predictive variation without excessive in-sample complexity.

Overall, Figure~\ref{fig:rmse_bar_all_datasets} supports the main empirical claim of this work: the proposed mixture-of-GAMs framework consistently improves upon standard global interpretable regression models across diverse benchmark problems, while remaining competitive with substantially more opaque alternatives. These results indicate that response-informed local adaptivity can provide a practical and principled compromise between predictive accuracy and model interpretability.

\subsubsection{Comparison with mixture-of-linear-model baselines}

We compare the proposed framework with the closely related locally adaptive baselines, namely MLM-cell and MLM-EPIC (Seo et al. \citeyear{MLM}), which combine DNN-cosupervised partitions with local linear predictors. Table~\ref{tab:mlm_compare} reports results on the California Housing and Bike Sharing datasets.

Across both benchmarks, the proposed mixture-of-GAMs method attains lower test RMSE than the MLM baselines. On the California Housing dataset, this improvement suggests that replacing local linear experts with cluster-specific additive smooth models is beneficial in a setting where the learned regimes are closely tied to spatially structured variation. The gain is also pronounced on the Bike Sharing dataset, where nonlinear temporal and meteorological effects are naturally accommodated within the specialized GAM components.

\begin{table}[htbp]
\centering
\caption{Comparison with mixture-of-linear-model (MLM) baselines on two benchmark datasets. Reported test uncertainties denote 95\% bootstrap confidence intervals. Lower RMSE indicates better predictive accuracy.}
\label{tab:mlm_compare}
\small
\setlength{\tabcolsep}{5pt}
\begin{tabular}{lcccc}
\toprule
& \multicolumn{2}{c}{California Housing} 
& \multicolumn{2}{c}{Bike Sharing} \\
\cmidrule(lr){2-3} \cmidrule(lr){4-5}
Method & Training RMSE & Test RMSE & Training RMSE & Test RMSE \\
\midrule
MLM-cell   & 0.560 & 0.570 $\pm$ 0.031 & 52.7 & 60.9 $\pm$ 7.1 \\
MLM-EPIC   & 0.569 & 0.584 $\pm$ 0.032 & 62.8 & 66.7 $\pm$ 6.9 \\
Mixture-of-GAMs & 0.442 & \textbf{0.489} $\pm$ 0.021 & 48.6 & \textbf{52.5} $\pm$ 2.4 \\
\bottomrule
\end{tabular}
\end{table}

\subsubsection{RFF-based data augmentation for sparse regression settings}

The preceding comparisons evaluate the proposed mixture-of-GAMs framework using the original training data. We further examine a data-augmentation variant in which the trained RFF model is used to generate responses for synthetic covariate points near the empirical data cloud. This strategy is analogous to the co-supervised construction of Seo et al.~(\citeyear{MLM}), where simulated covariate points are assigned responses by a trained DNN and then used to estimate local models. In the present work, the fitted RFF model plays the corresponding role of the high-accuracy reference model.

Starting from the standardized training data, we generate synthetic covariate candidates by adding Gaussian perturbations to observed samples and retain only those candidates that remain close to the empirical data cloud according to a Mahalanobis-distance criterion (Murphy \citeyear{Murphy_book}, Chapter~3.2). The retained covariates are assigned RFF-generated responses using the trained random feature regression model and are combined with the observed training sample, whose covariates are kept on the same standardized scale. The mixture-of-GAMs predictor is subsequently retrained on this augmented sample.

As summarized in Table~\ref{tab:data_aug}, this augmentation strategy improves predictive performance in the two augmentation experiments, conducted on the Airfoil Self-Noise and Kin40k datasets. Relative to the corresponding baseline mixture-of-GAMs model trained on the observed, non-augmented data, the test RMSE is reduced by approximately $9\%$ on the Airfoil Self-Noise dataset and $5\%$ on the Kin40k dataset. For Kin40k, uncertainty is estimated using the residual-bootstrap procedure described in Section~\ref{subsec_numerical_setup}. For the smaller Airfoil Self-Noise dataset, we use $100$ repeated random training/test splits, also known as Monte Carlo cross-validation (Kuhn and Johnson, \citeyear{Applied_pred}, Chapter~4), and report the mean RMSE with $95\%$ confidence intervals computed as $1.96$ standard errors across repetitions. These reductions suggest that the response-informed spectral model can provide useful pseudo-responses for improving coverage in sparsely sampled regions of the covariate space, thereby supporting the downstream local GAM fitting.

\begin{table}[htbp]
\centering
\renewcommand{\arraystretch}{1.2}
\caption{Impact of RFF-guided data augmentation on predictive performance. Reported values are test RMSE with uncertainty estimates computed as described in the text.}
\label{tab:data_aug}
\begin{tabular}{lccc}
\toprule
{Dataset} 
& \makecell{\footnotesize{RMSE of Base}\\\footnotesize{Mixture-of-GAMs}} 
& \makecell{\footnotesize{RMSE of Augmented}\\\footnotesize{Mixture-of-GAMs}} 
& \makecell{\footnotesize{Relative}\\\footnotesize{Improvement}} \\
\midrule
{Airfoil Self-Noise} 
& $2.22 \pm 0.05$ 
& $2.02 \pm 0.04$
& $\approx 9\%$ \\
{Kin40k} 
& $0.339 \pm 0.012$ 
& $0.324 \pm 0.011$ 
& $\approx 5\%$ \\
\bottomrule
\end{tabular}
\end{table}

\subsection{Representation-guided regime discovery and interpretability}
\label{subsection_regime}
Beyond predictive accuracy, an important objective of the proposed framework is to determine whether the learned latent regimes correspond to meaningful structure in the underlying application domain. Because the clustering stage is performed on response-informed representations rather than directly on the raw covariates, the resulting partitions need not reflect simple geometric proximity in the original input space. We therefore examine whether the learned responsibilities reveal interpretable spatial or temporal organization in two representative datasets.

\paragraph{Spatial regime discovery in California Housing dataset.}
We first study the California Housing dataset, where geographic heterogeneity is anticipated to be an important driver of predictive variation. Let $(\xi_i,\eta_i)$ denote the spatial coordinates obtained from the longitude and latitude of the $i$-th observation. Restricting the RFF representation to these two variables yields a spatial-RFF model with frequency vectors
\[
\boldsymbol{\omega}_k=(\omega_{k,\xi},\omega_{k,\eta})\in\mathbb{R}^2,\qquad k=1,\dots,K,
\]
so that the resulting feature representation captures oscillatory variation directly over geographic coordinates. Although this simplified model uses only two covariates, it still attains a competitive test RMSE of approximately $0.56\times 10^5$ USD, indicating that spatial location alone contains substantial predictive information.

\begin{figure}[htbp]
  \centering

  \begin{subfigure}[b]{0.24\textwidth}
    \includegraphics[width=\textwidth]{ 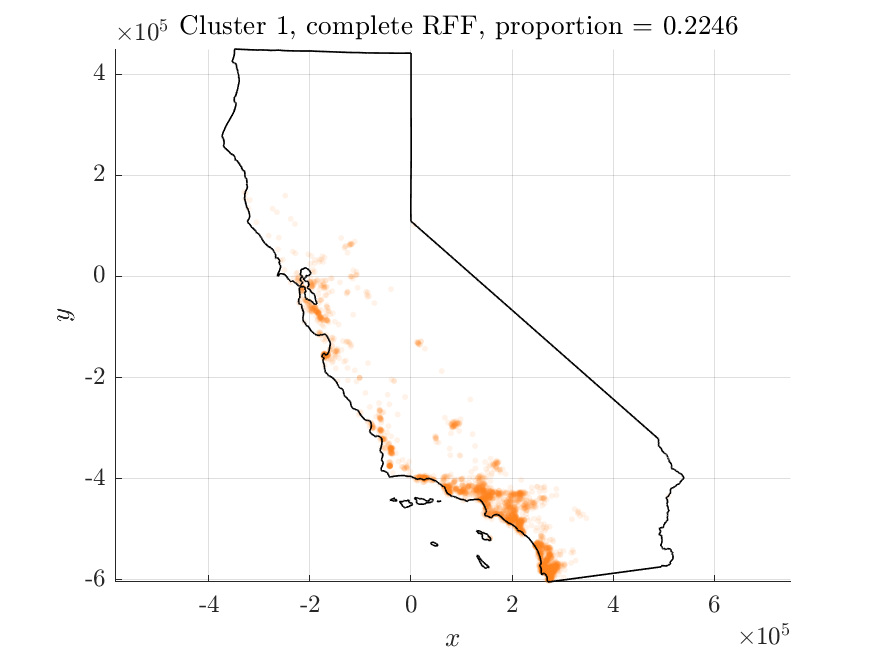}
    \caption{Cluster 1}
  \end{subfigure}
  \begin{subfigure}[b]{0.24\textwidth}
    \includegraphics[width=\textwidth]{ 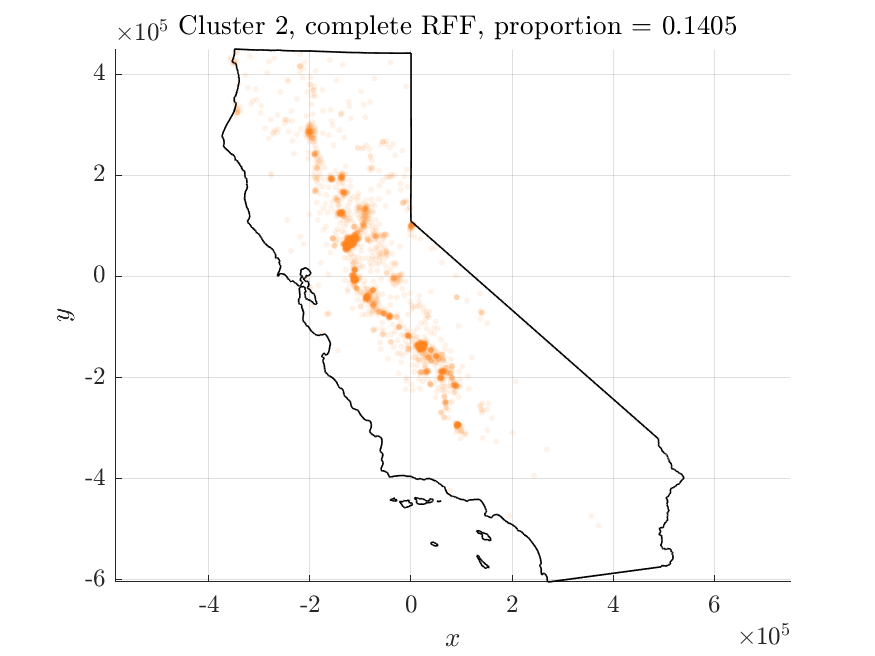}
    \caption{Cluster 2}
  \end{subfigure}
  \begin{subfigure}[b]{0.24\textwidth}
    \includegraphics[width=\textwidth]{ 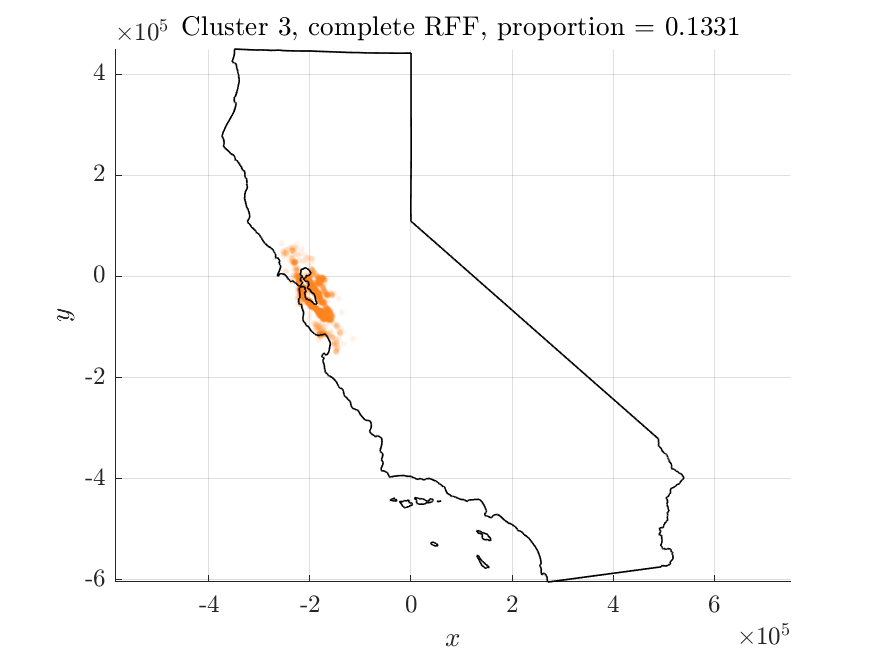}
    \caption{Cluster 3}
  \end{subfigure}
  \begin{subfigure}[b]{0.24\textwidth}
    \includegraphics[width=\textwidth]{ 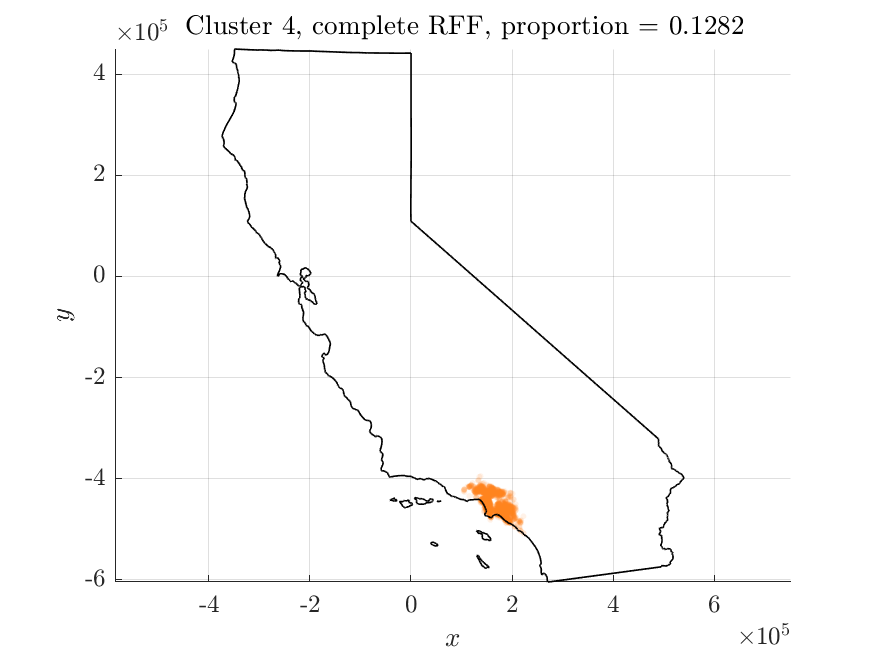}
    \caption{Cluster 4}
  \end{subfigure}
  
  \begin{subfigure}[b]{0.24\textwidth}
    \includegraphics[width=\textwidth]{ 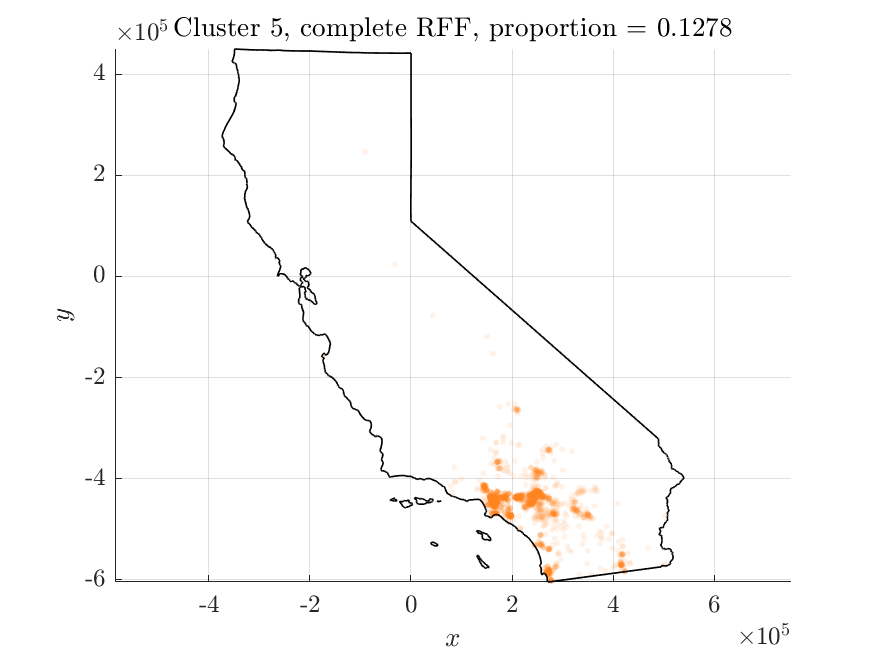}
    \caption{Cluster 5}
  \end{subfigure}
  \begin{subfigure}[b]{0.24\textwidth}
    \includegraphics[width=\textwidth]{ 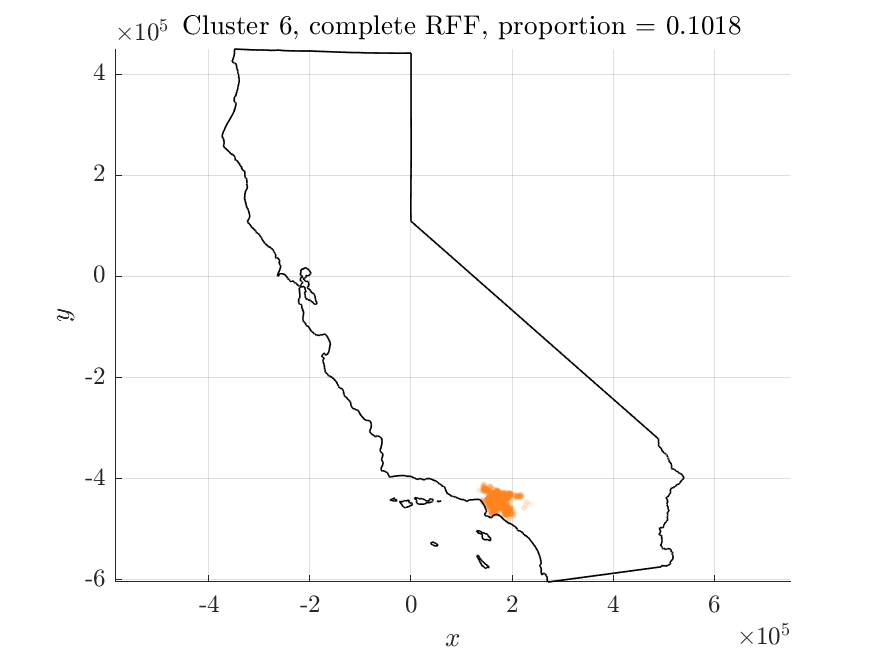}
    \caption{Cluster 6}
  \end{subfigure}
  \begin{subfigure}[b]{0.24\textwidth}
    \includegraphics[width=\textwidth]{ 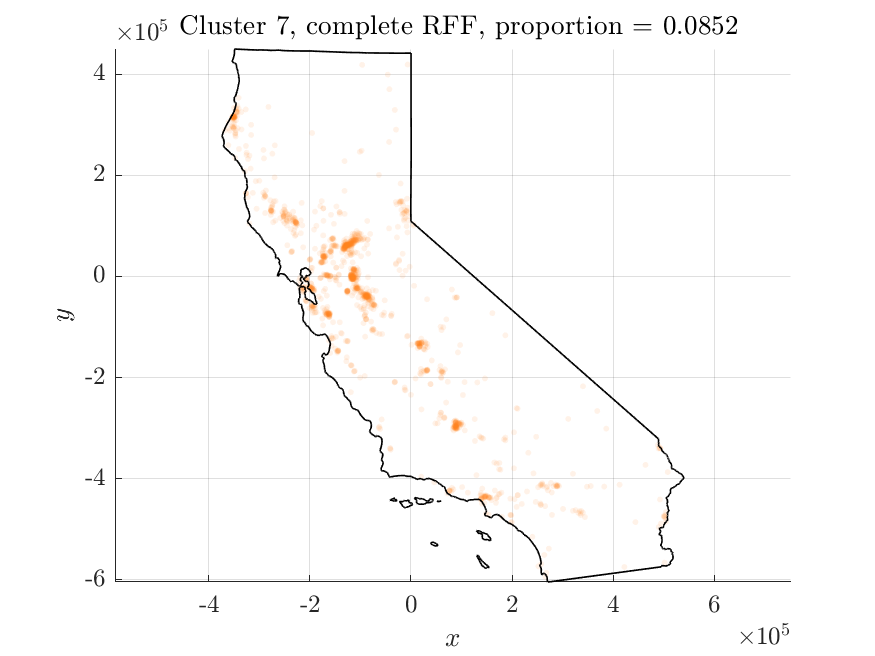}
    \caption{Cluster 7}
  \end{subfigure}
  \begin{subfigure}[b]{0.24\textwidth}
    \includegraphics[width=\textwidth]{ 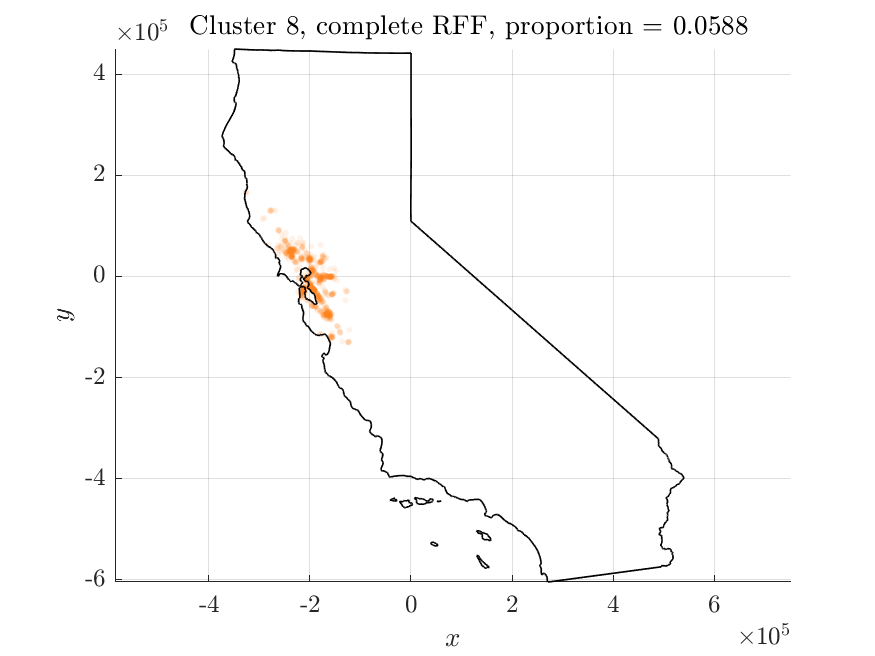}
    \caption{Cluster 8}
  \end{subfigure}

   \caption{Spatial distributions of training data for each GMM cluster in the California housing dataset, using complete random Fourier features.}
  \label{fig:clusters_subfig_comp}
\end{figure}

\begin{figure}[htbp]
  \centering

  \begin{subfigure}[b]{0.24\textwidth}
    \includegraphics[width=\textwidth]{ 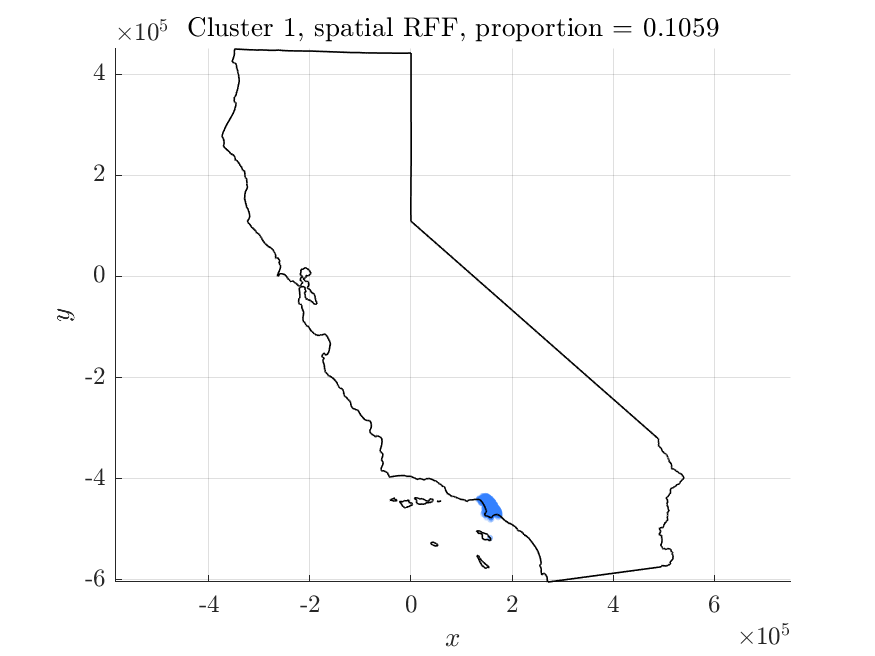}
    \caption{Cluster 1}
  \end{subfigure}
  \begin{subfigure}[b]{0.24\textwidth}
    \includegraphics[width=\textwidth]{ 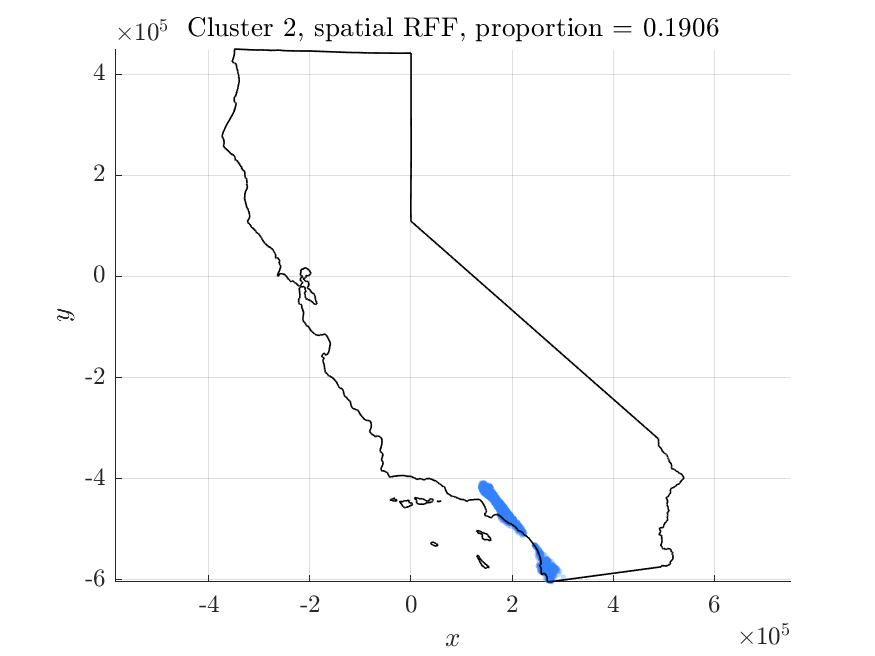}
    \caption{Cluster 2}
  \end{subfigure}
  \begin{subfigure}[b]{0.24\textwidth}
    \includegraphics[width=\textwidth]{ 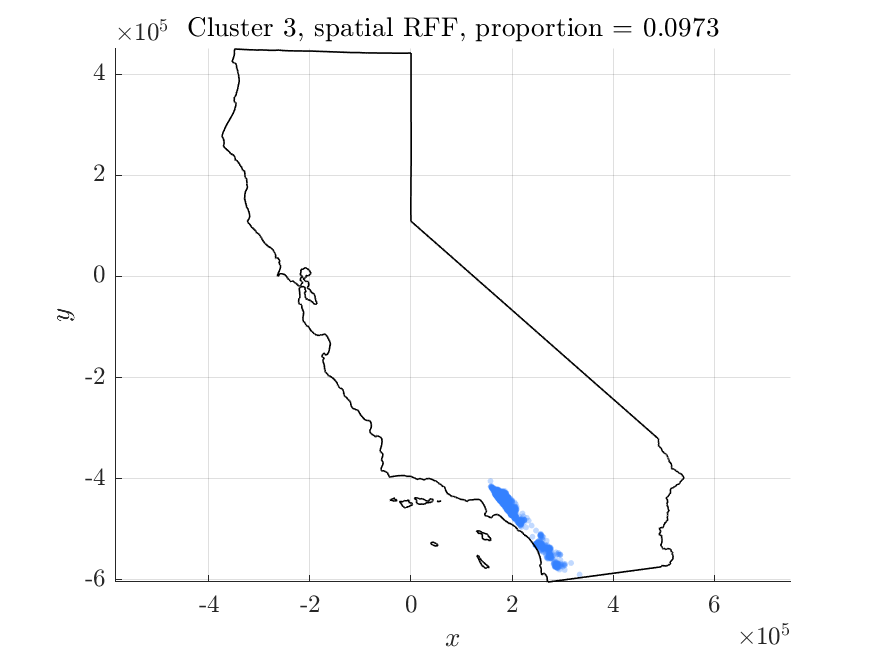}
    \caption{Cluster 3}
  \end{subfigure}
  \begin{subfigure}[b]{0.24\textwidth}
    \includegraphics[width=\textwidth]{ 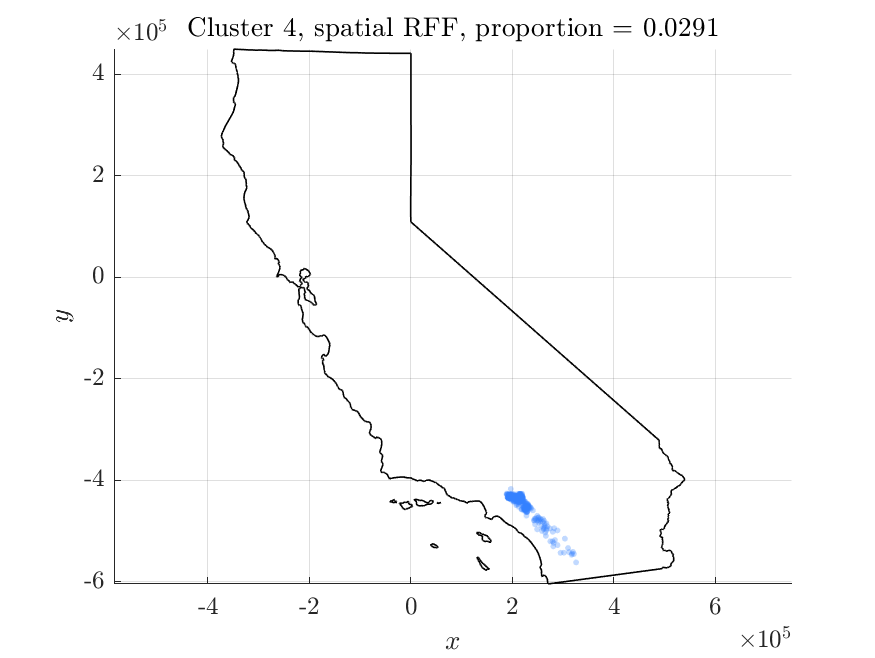}
    \caption{Cluster 4}
  \end{subfigure}
  
  \begin{subfigure}[b]{0.24\textwidth}
    \includegraphics[width=\textwidth]{ 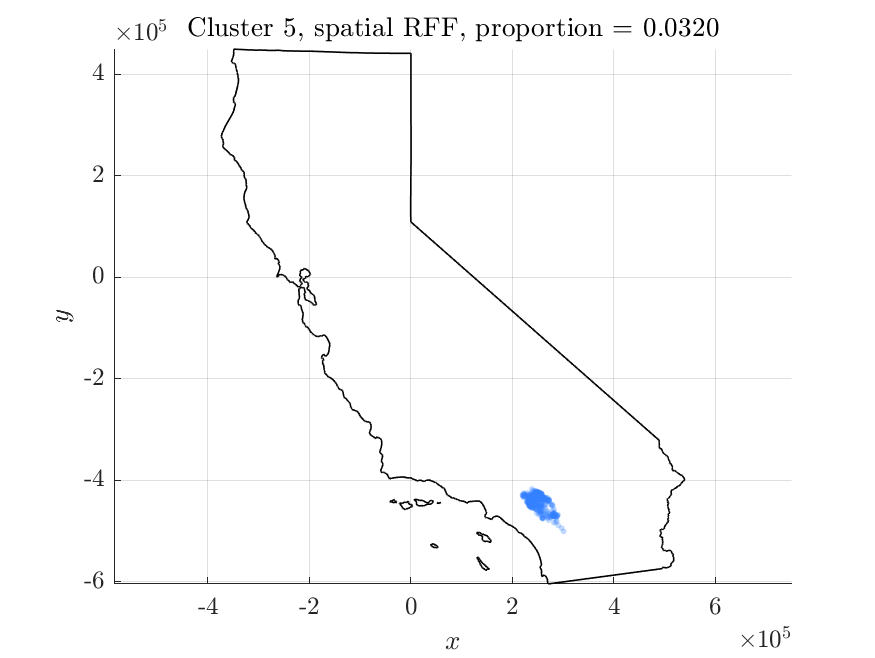}
    \caption{Cluster 5}
  \end{subfigure}
  \begin{subfigure}[b]{0.24\textwidth}
    \includegraphics[width=\textwidth]{ 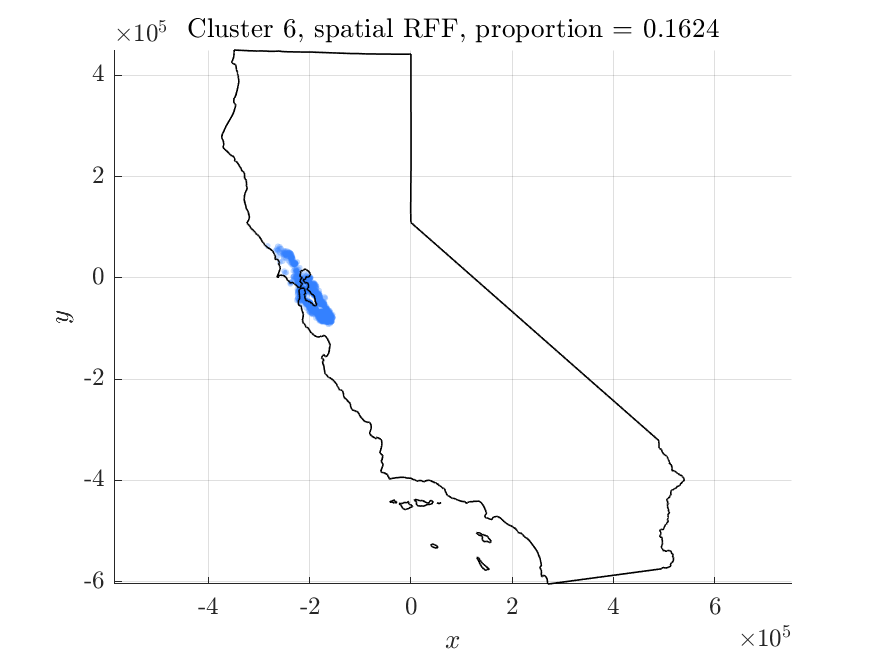}
    \caption{Cluster 6}
  \end{subfigure}
  \begin{subfigure}[b]{0.24\textwidth}
    \includegraphics[width=\textwidth]{ 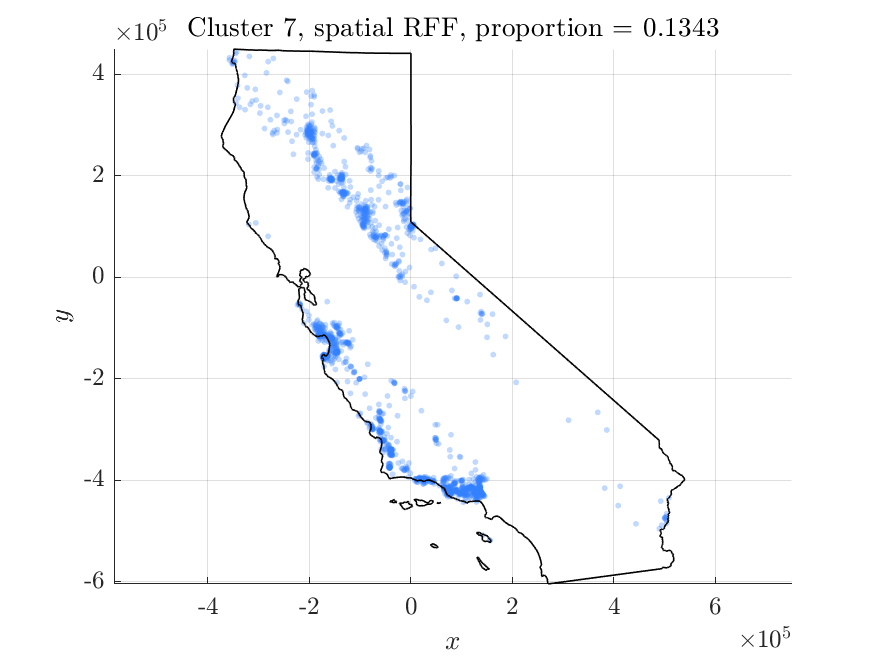}
    \caption{Cluster 7}
  \end{subfigure}
  \begin{subfigure}[b]{0.24\textwidth}
    \includegraphics[width=\textwidth]{ 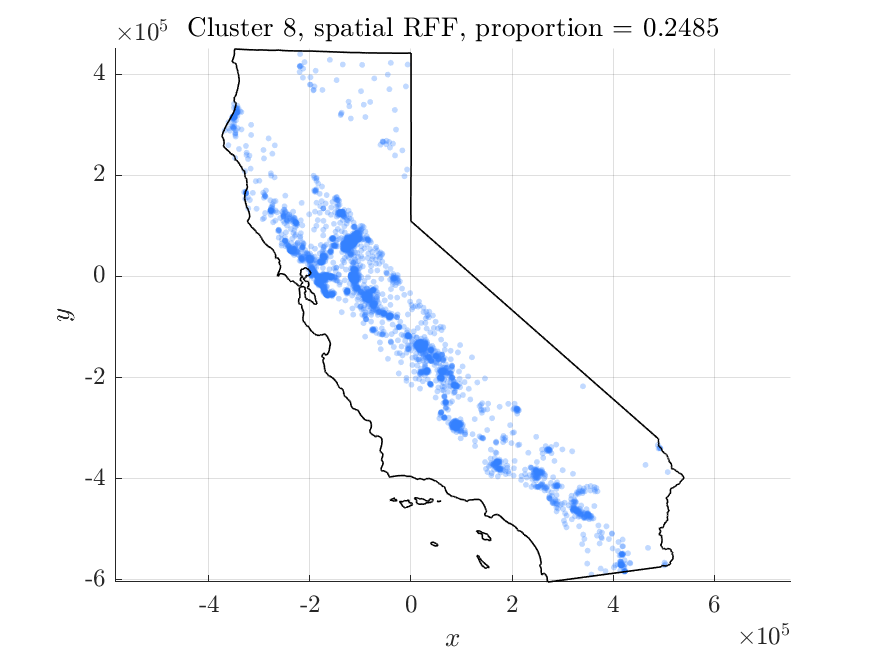}
    \caption{Cluster 8}
  \end{subfigure}

   \caption{Spatial distributions of training data for each GMM cluster in the California housing dataset, based on explainable spatial Fourier features.}
  \label{fig:clusters_subfig_spat}
\end{figure}

Figures~\ref{fig:clusters_subfig_comp} and~\ref{fig:clusters_subfig_spat} compare the spatial distributions of the learned GMM components under two constructions: a complete-RFF representation based on all covariates, and a spatial-RFF representation based only on geographic coordinates. In both cases, the learned regimes exhibit clear spatial coherence, with several components concentrating around recognizable regions such as major coastal population centers and inland areas associated with the Central Valley. This suggests that the response-informed representation identifies meaningful regional structure rather than arbitrary partitions. At the same time, the two specifications exhibit a notable difference. Under the complete-RFF representation using eight covariates, the clusters remain spatially organized but are relatively localized and irregular, reflecting the influence of the full set of predictive variables. By contrast, the spatial-RFF construction produces a cleaner large-scale partition, with several components forming elongated bands that are more parallel to the California coastline. This makes the learned regime structure substantially easier to interpret in geographic terms and suggests that variation in distance from the coastline, rather than movement along the coastline itself, is a primary driver of the partition. This geographic pattern is examined in greater detail in the California Housing case study of Section~\ref{Subsection_Cal}.

\paragraph{Temporal regime discovery in the Bike Sharing dataset.}

We next consider the Bike Sharing dataset, where heterogeneity is driven primarily by recurring temporal demand cycles rather than by spatial variation. Rental activity is shaped by periodic usage patterns, including commuting peaks, leisure periods, and seasonal effects, while continuous meteorological variables contribute to smoother nonlinear variation. This combination makes the dataset a natural setting for examining whether the proposed framework can recover meaningful operating regimes from the predictive structure.

To visualize the learned soft partition, we examine how the mixture components are activated across temporal contexts. Let $\gamma_{i\ell}$ denote the posterior responsibility of mixture component $\ell$ for observation $i$, and let $h_i\in\{0,1,\dots,23\}$ denote the associated hour of day. We summarize the average responsibility at hour $\tau$ by
\[
\bar{\gamma}_{\ell}(\tau)
=
\frac{1}{|\mathcal I_\tau|}
\sum_{i\in\mathcal I_\tau}\gamma_{i\ell},
\qquad
\mathcal I_\tau:=\{i:h_i=\tau\}.
\]

\begin{figure}[htbp]
    \centering
    \includegraphics[width=1.0\linewidth]{ 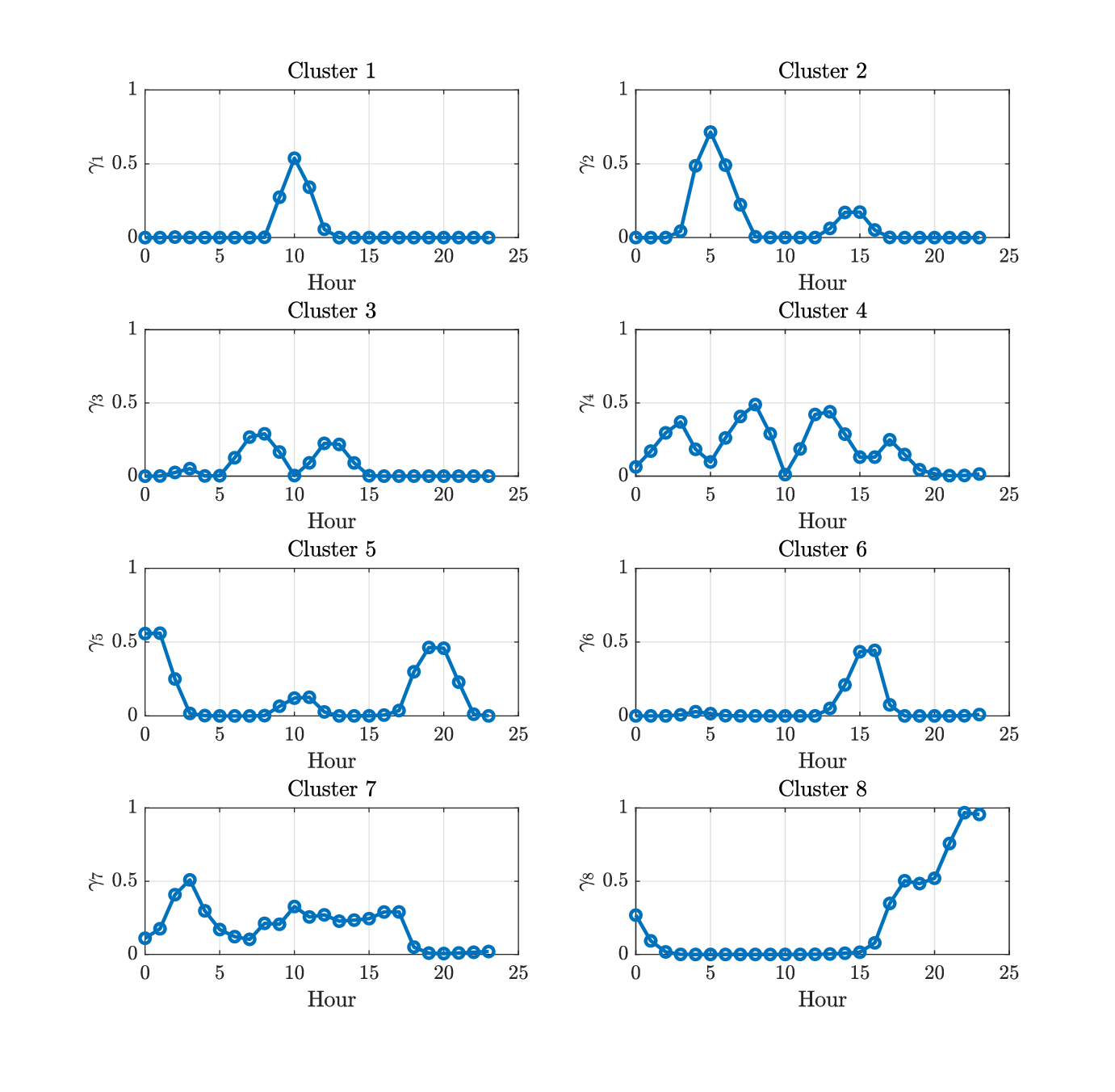}
    \vspace{-1.2cm}
    \caption{Average posterior responsibilities $\bar{\gamma}_\ell(\tau)$ of the eight mixture components as functions of the hour of day, computed over the training dataset.}
    \label{fig:bike_hour_resp}
\end{figure}

Figure~\ref{fig:bike_hour_resp} plots $\bar{\gamma}_{\ell}(\tau)$ for the learned mixture components. Distinct temporal regimes emerge clearly: several components exhibit pronounced activation during the morning and evening hours, whereas others are more active during midday or late-evening periods. These patterns align naturally with well-known bike-sharing usage behavior, including commuting-related demand during peak hours and more flexible leisure-oriented activity outside those intervals. These findings indicate that the proposed framework is able to infer meaningful temporal operating regimes directly from the learned response-informed representation, without requiring manually imposed time-of-day segmentation.

\subsection{Case studies and diagnostics}\label{Subsection_Cal}

\subsubsection{California Housing: representation and geographic regimes}
The California Housing dataset provides a particularly informative case study for the proposed framework because it combines strong predictive heterogeneity with a spatial structure that admits direct interpretation. In addition to serving as a benchmark for predictive accuracy, this dataset enables examination of how the learned random Fourier feature representation, the induced latent partition, and the resulting local GAM components relate to meaningful large-scale geographic variation.

\paragraph{Stability of the learned spectral representation.}

We begin by examining the efficacy of the resampling-based random Fourier feature model, since the subsequent clustering and frequency-based interpretation rely on this learned spectral representation. The model is trained using the resampling scheme proposed by Huang et al.~(\citeyear{RFF_resampling}), which aims to improve the efficiency of frequency selection by leveraging the empirical covariance structure of the sampled frequencies. The number of random Fourier features $K$ is selected by gradually increasing its value until validation performance stabilizes, indicating that the approximation error induced by a finite feature budget is no longer the dominant source of error. For this dataset, the final model uses $K=4000$ sampled frequencies.

\begin{figure}[htbp]
    \centering
    \includegraphics[width=0.9\linewidth]{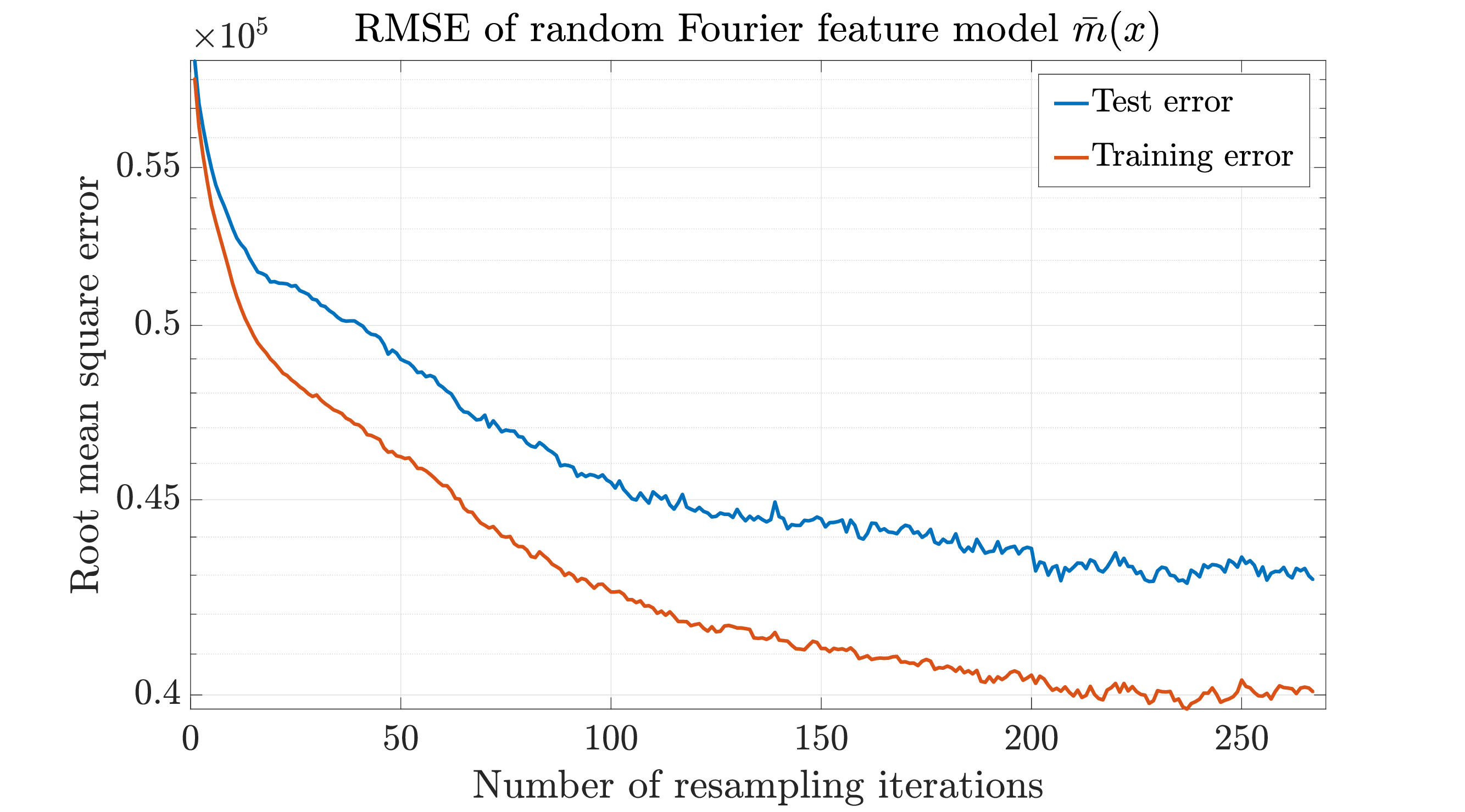}
    \caption{Root mean square error of the trained random Fourier feature model $\bar{m}(\bmx)$ as the number of frequency-resampling iterations increases.}
    \label{fig:RFF_resampling_error_curve}
\end{figure}

Figure~\ref{fig:RFF_resampling_error_curve} reports the training and test RMSE values over successive resampling iterations. Both curves decrease markedly in the early iterations and then continue to improve at a slower rate before stabilizing at later stages, indicating convergence of the resampling procedure. The close tracking of the test curve with no substantial late-stage deterioration suggests that the learned representation remains stable throughout the iterations. These results support using the resulting spectral model as the basis for the subsequent analysis of learned frequencies and latent regime structure.

\paragraph{Random feature-guided frequency orientation and spatial structure.}
Section~\ref{subsection_regime} showed that the spatial-RFF construction yields geographically coherent mixture components, with several latent regimes forming elongated bands approximately parallel to the California coastline. We proceed to characterize the broader spatial structure underlying this partition by analyzing the learned distribution of spatial frequencies.

\begin{figure}[htbp]
    \centering
    \begin{subfigure}[b]{0.49\textwidth}
        \includegraphics[width=0.9\linewidth]{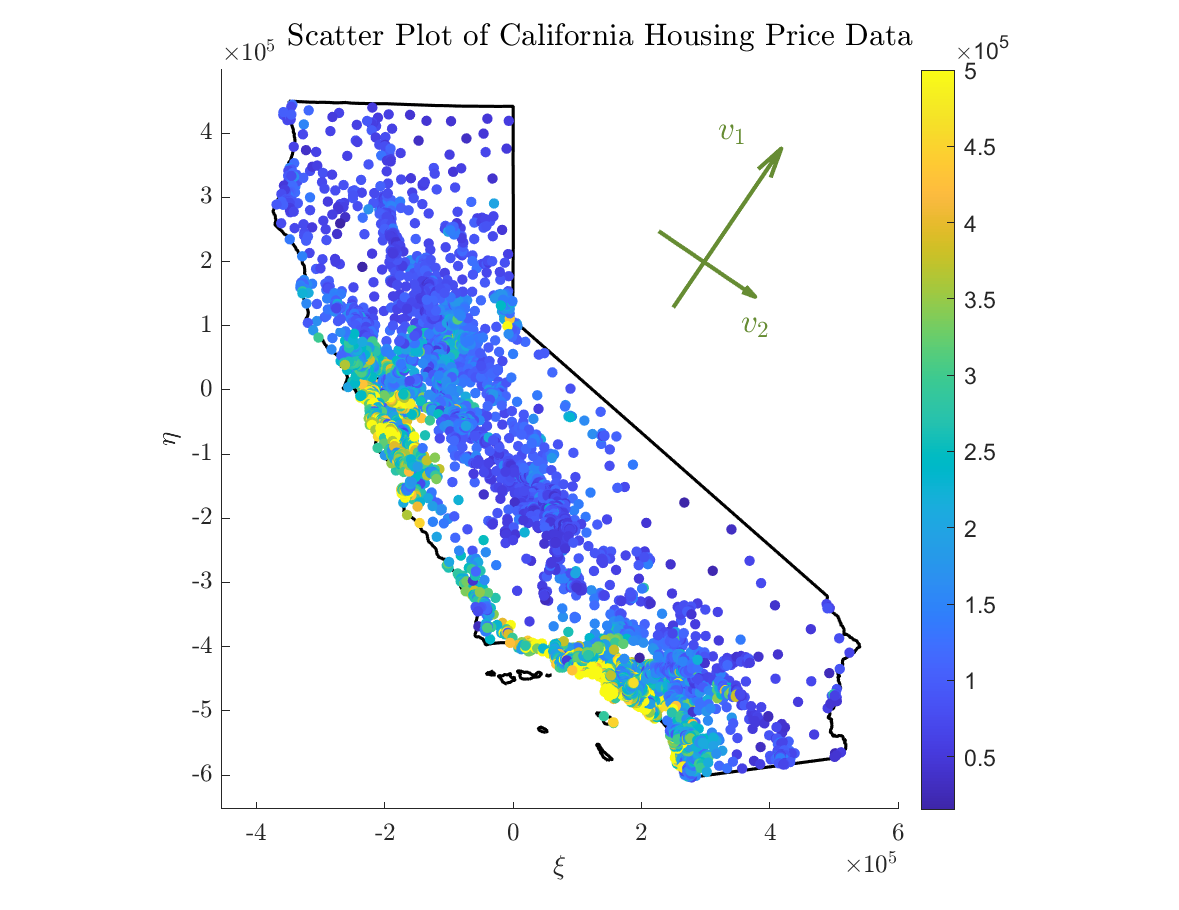}
        \caption{California housing price data.}
        \label{fig:RFF_explain_a}
    \end{subfigure}
    \begin{subfigure}[b]{0.49\textwidth}
        \includegraphics[width=0.9\linewidth]{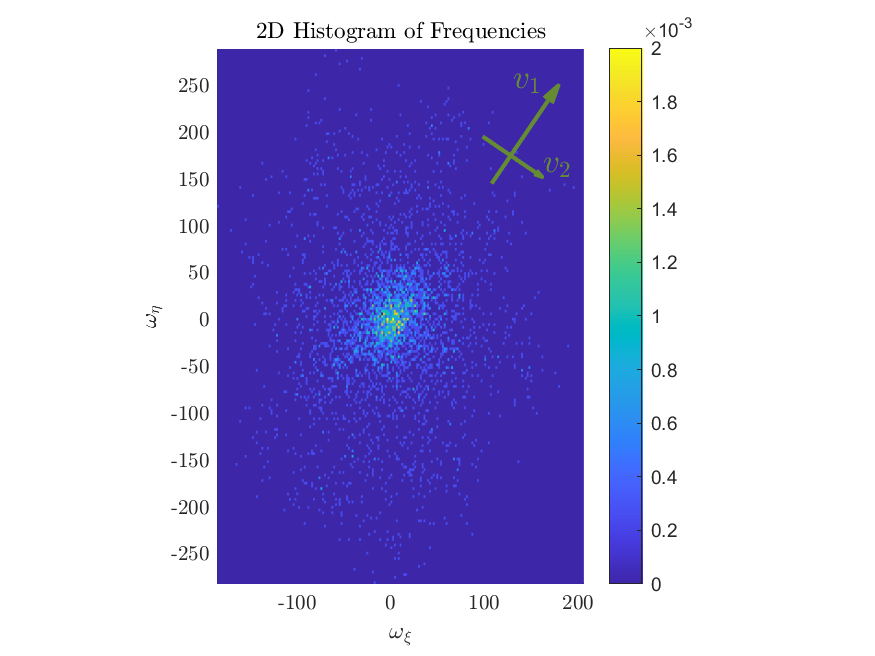}
        \caption{2D histogram of spatial frequencies.}
        \label{fig:RFF_explain_b}
    \end{subfigure}
    \caption{California housing price data (left) and empirical histogram of the learned spatial frequency samples $\{(\omega_{k,\xi},\omega_{k,\eta})\}_{k=1}^K$ (right). In both panels, $v_1$ denotes the dominant direction identified by the weighted PCA of the frequency distribution, and $v_2$ denotes the orthogonal direction.}
    \label{fig:RFF_explain}
\end{figure}

To investigate this structure, we visualize the empirical histogram of the two-dimensional frequency samples $\{(\omega_{k,\xi},\,\omega_{k,\eta})\}_{k=1}^K$ in Figure~\ref{fig:RFF_explain_b}. Geographically, California's coastline extends from the northwest to the southeast, while the orthogonal inland direction (which roughly points from southwest to northeast) coincides with a systematic decline in median house values as proximity to coastal urban centers decreases, as illustrated in Figure~\ref{fig:RFF_explain_a}.
To quantify the dominant orientation within the learned spectral distribution, we apply a weighted PCA based on kernel density estimates computed using the classical Rosenblatt-Parzen formulation (Rosenblatt \citeyear{Rosenblatt_1956}; Parzen \citeyear{Parzen_1962}), which emphasizes the high-density core of the frequency distribution. The resulting principal direction (denoted by $v_1$ in Figures~\ref{fig:RFF_explain_a} and \ref{fig:RFF_explain_b}) aligns closely with the coastal-inland gradient observed in the housing price data. This alignment suggests that the spatial RFF model, bolstered by the adaptive resampling method, effectively captures the primary axis of geographic variation and extracts meaningful large-scale structural features from the data.

\begin{figure}[htbp]
    \centering
    \begin{subfigure}[b]{0.45\textwidth}
        \includegraphics[width=0.9\linewidth]{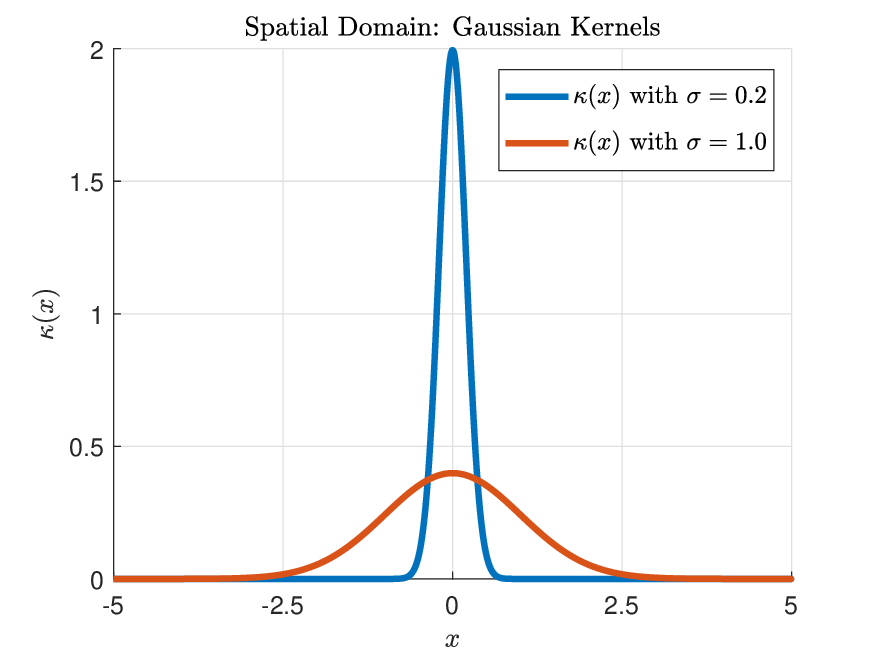}
        \caption{Kernel function $\kappa(x)$ in the spatial domain.}
    \end{subfigure}
    \begin{subfigure}[b]{0.45\textwidth}
        \includegraphics[width=0.9\linewidth]{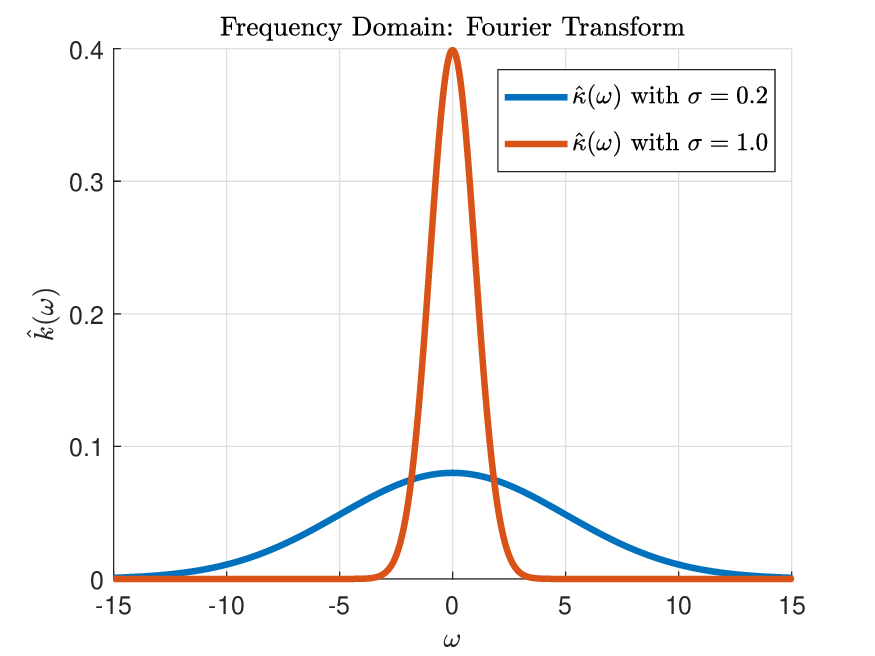}
        \caption{Fourier transform $\hat{\kappa}(\omega)$ in the frequency domain.}
    \end{subfigure}
    \caption{Illustration of the inverse relation between the spatial and spectral scales of Gaussian kernels. Left: Gaussian kernels with different bandwidth parameters $\sigma$ in the spatial domain. Right: the corresponding normalized Fourier transforms, showing broader spectral spread for smaller $\sigma$.}
    \label{fig:spatial_freq_tradeoff}
\end{figure}

This phenomenon can be understood through the duality between spatial smoothness and spectral decay. Directional variation in the target function dictates the required kernel scale: rapid fluctuations along the coastal-inland axis necessitate sharper kernels with smaller spatial bandwidths. Given that the Fourier transform of the Gaussian RBF kernel $\kappa_\sigma$ in \eqref{eq:Gaussian_RBF} exhibits a spectral variance inversely proportional to the spatial variance $\sigma^2$, these localized spatial variations yield a broader distribution of frequency components along the corresponding axis. By contrast, directions characterized by smoother variations require less spectral support and remain concentrated near the low-frequency region.
Figure~\ref{fig:spatial_freq_tradeoff} illustrates this inverse relation between spatial scale and spectral spread. This rationale is consistent with the anisotropy observed in the learned frequency samples.

We also assess whether the learned spatial representation is sufficient for downstream clustering. Using only the spatial Fourier features $\{(\omega_{k,\xi},\omega_{k,\eta})\}_{k=1}^K$ and the associated amplitudes $\{\beta_k\}_{k=1}^K$, we perform GMM-based clustering on the resulting latent representation and fit the corresponding cluster-wise GAM models. This variant attains a test RMSE of $0.489\times10^5$ USD, which is slightly smaller than the $0.501\times10^5$ USD obtained when clustering is guided by the full random Fourier feature representation over all eight covariates. These results suggest that the dominant regime structure in this dataset is largely captured by the spatial component alone.

\begin{figure}[htbp]
    \centering
    \begin{subfigure}[b]{0.32\textwidth}
        \includegraphics[width=\textwidth]{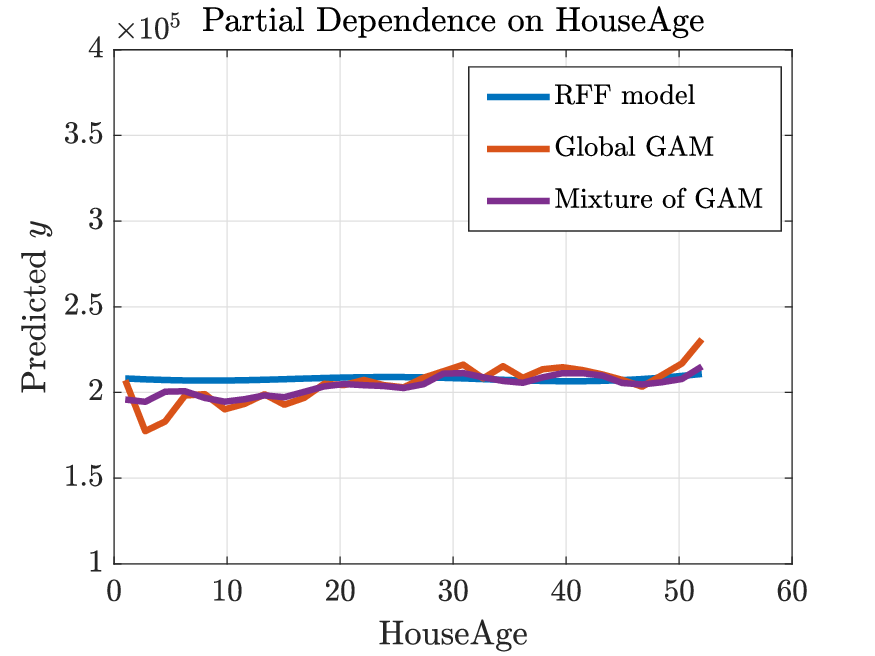}
        \caption{Median house age}
    \end{subfigure}
    \hfill
    \begin{subfigure}[b]{0.32\textwidth}
        \includegraphics[width=\textwidth]{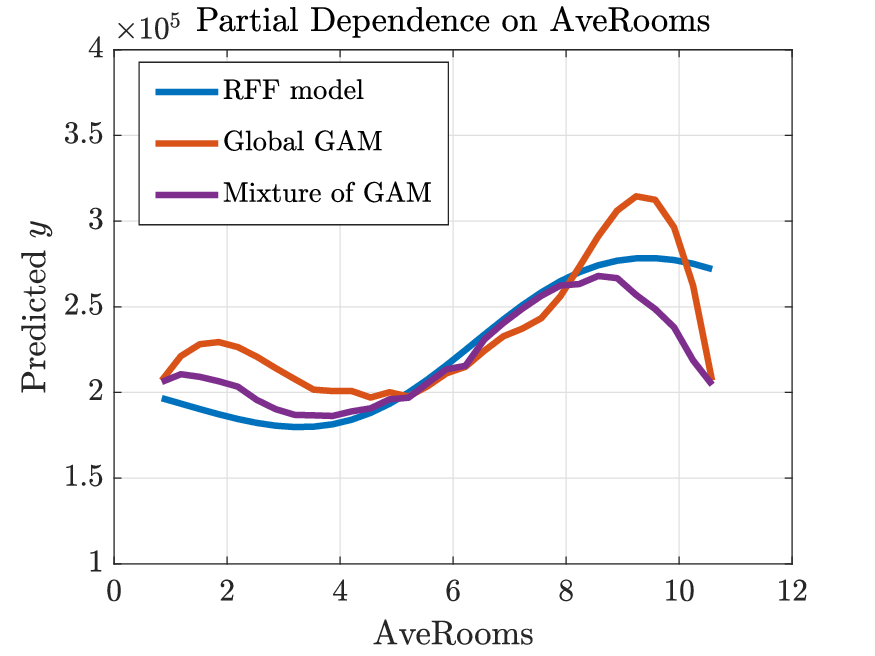}
        \caption{Avg. no. of rooms}
    \end{subfigure}
    \hfill
    \begin{subfigure}[b]{0.32\textwidth}
        \includegraphics[width=\textwidth]{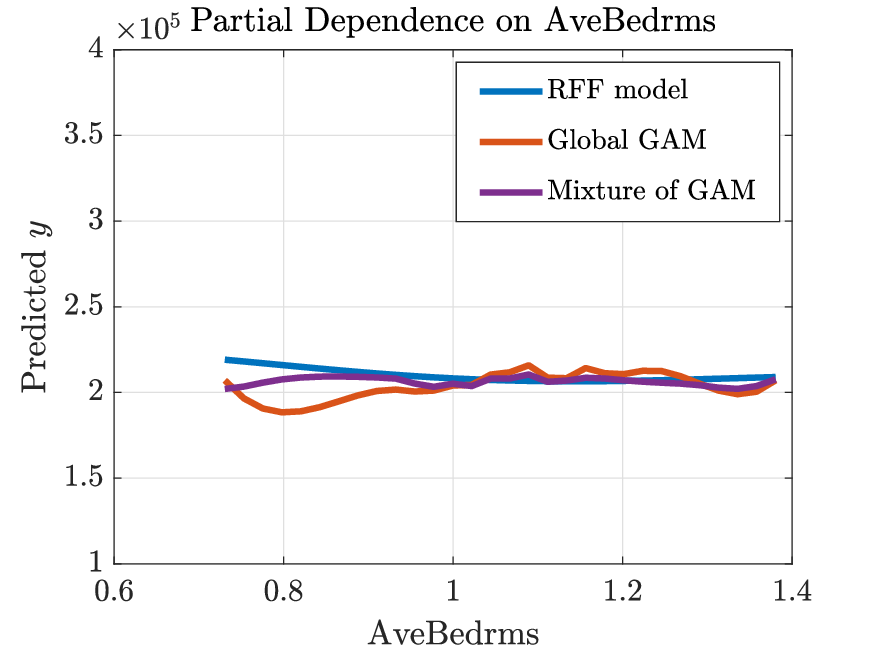}
        \caption{Avg. no. of bedrooms}
    \end{subfigure}

    \vspace{0.5cm}

    \begin{subfigure}[b]{0.32\textwidth}
        \includegraphics[width=\textwidth]{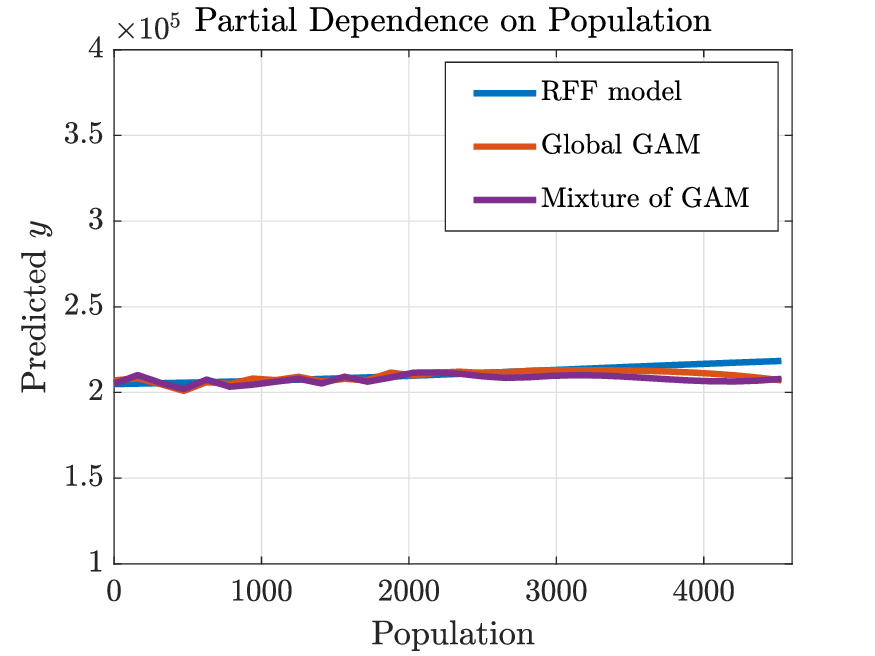}
        \caption{Block population}
    \end{subfigure}
    \hfill
    \begin{subfigure}[b]{0.32\textwidth}
        \includegraphics[width=\textwidth]{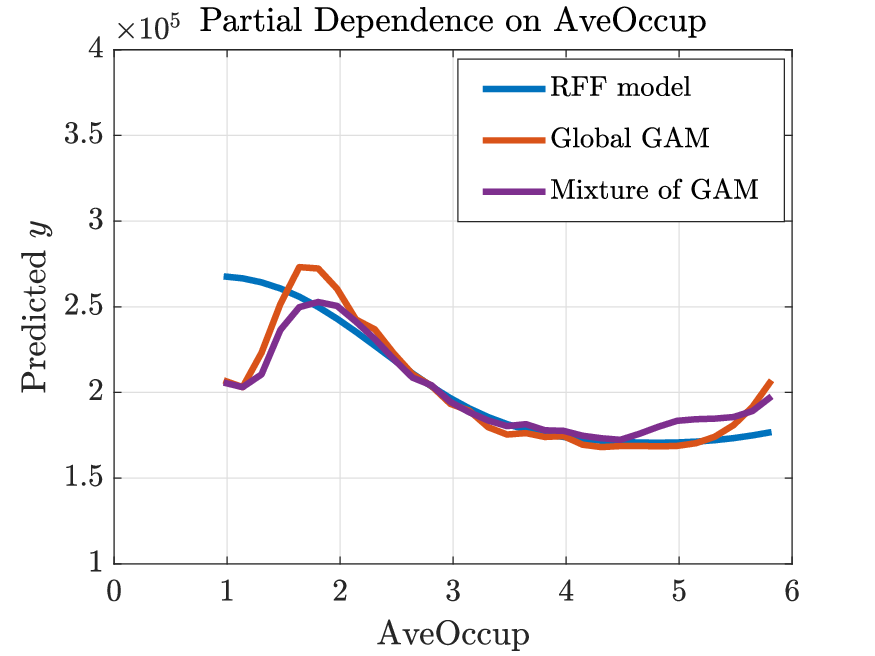}
        \caption{Avg. occupancy}
    \end{subfigure}
    \hfill
    \begin{subfigure}[b]{0.32\textwidth}
        \includegraphics[width=\textwidth]{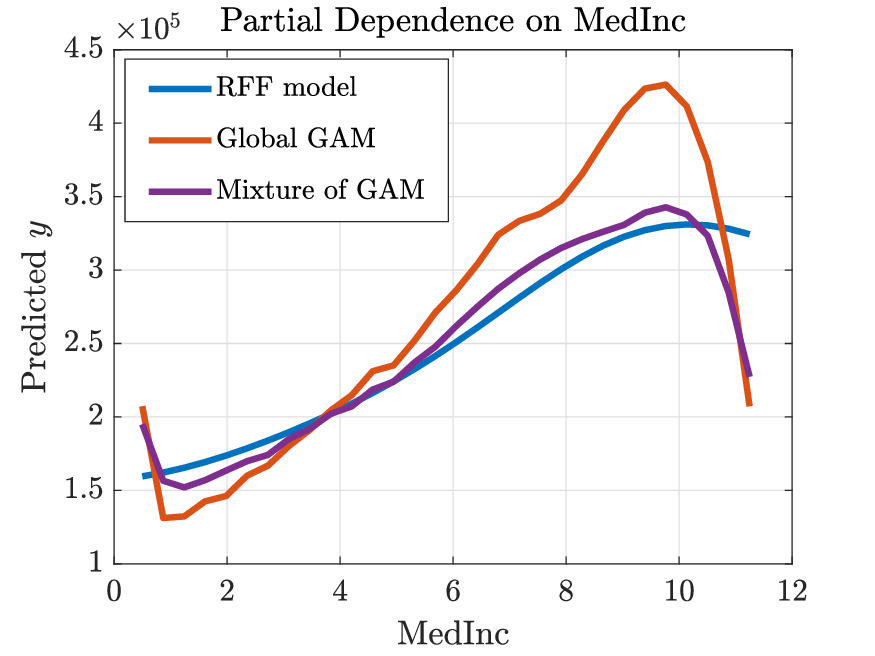}
        \caption{Median income}
    \end{subfigure}

    \caption{Partial dependence plots for selected features of the California Housing dataset, comparing the RFF model, the global GAM, and the proposed mixture-of-GAMs.}
    \label{fig:partial_dependence}
\end{figure}

\paragraph{Comparison of covariate effects through partial dependence.}
We further compare the fitted covariate-response relationships through partial dependence plots (Friedman \citeyear{Friedman_pdp}). For a covariate $x_j$, the partial dependence function is defined by
\[
\mathrm{PD}_j(x_j)
=
\frac{1}{N}\sum_{i=1}^N \bar{m}(x_j,\bmx_{i,-j}),
\]
where $\bmx_{i,-j}\in\mathbb{R}^{p-1}$ denotes the vector of all remaining covariates from the $i$-th training sample. Figure~\ref{fig:partial_dependence} displays the resulting curves for six selected variables, comparing the mixture of GAMs with the RFF model and the global GAM.

Across these covariates, the mixture of GAMs generally follows the dominant patterns of the RFF reference more closely than the single global GAM. This is particularly visible for median income, house age, and average number of bedrooms, where the local additive formulation captures nonlinear changes that are smoothed out under the global model. These plots complement the spatial analysis above: beyond identifying meaningful latent regimes, the proposed framework also recovers interpretable covariate-response patterns that more closely reflect the nonlinear structure learned by the underlying spectral model.

\subsubsection{Airfoil Self-Noise: modeling coupled physical mechanisms under data scarcity}

While the California Housing study highlights geographic interpretability, the Airfoil Self-Noise dataset provides a complementary evaluation in a setting characterized by complex physical interactions and limited data. Derived from NASA wind-tunnel experiments, this task requires predicting sound pressure levels based on covariates such as acoustic frequency, angle of attack, and displacement thickness. These variables are known to exhibit strong nonlinear coupling, and the relatively sparse coverage of the covariate space increases the risk of overfitting for highly parameterized models.

\paragraph{Establishing a high-fidelity spectral reference.}
As established in the broader performance comparison of Section~\ref{sec:predictive_performance_overall}, the resampling-based RFF model serves as a strong predictive reference, attaining a test RMSE of $1.08 \pm 0.02$ dB. This performance not only exceeds that of tree-based ensembles like XGBoost but also matches recent state-of-the-art result (test RMSE around $1.2$ dB) achieved by a hybrid CatBoost-AOA approach (Rastgoo and Khajavi \citeyear{Rastgoo2023}). For the proposed framework, this high-fidelity model is critical, as it provides the response-informed representation that guides the identification of local predictive regions. The mixture-of-GAMs framework substantially improves upon the global GAM (test RMSE of 4.5 dB), achieving an accuracy of 2.2 dB that is statistically competitive with that of Explainable Boosting Machine, as depicted in the second panel of Figure~\ref{fig:rmse_bar_all_datasets}. This performance is primarily driven by the RFF-based clustering to uncover latent data regimes, enabling local specialization to capture aerodynamic mechanisms that are otherwise obscured by global averaging.

\paragraph{Analyzing model behavior via partial dependence.}

\begin{figure}[ht!]
\centering

\begin{subfigure}[b]{0.32\textwidth}
    \includegraphics[width=\textwidth]{ 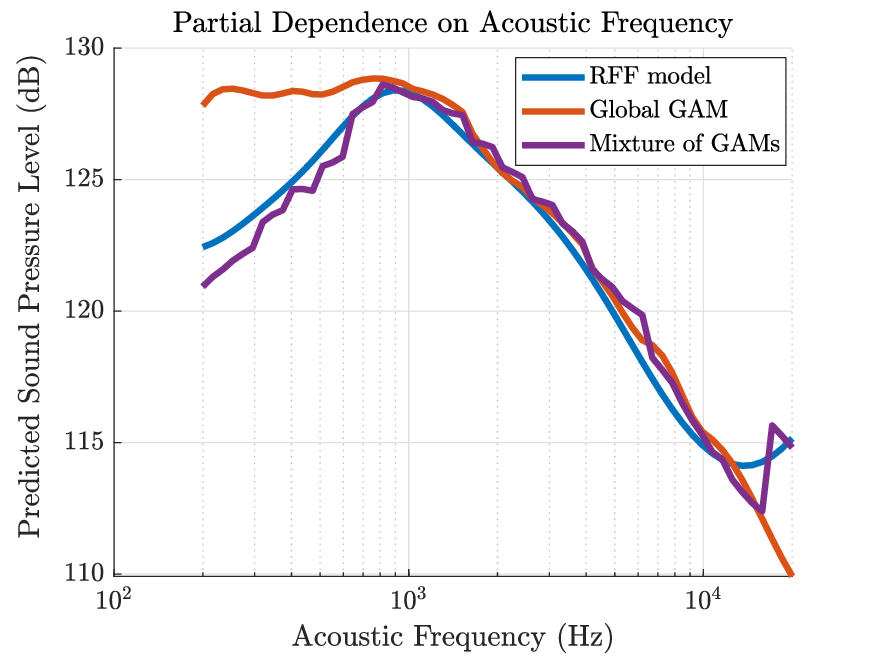}
    \caption{Acoustic Frequency}
\end{subfigure}
\hfill
\begin{subfigure}[b]{0.32\textwidth}
    \includegraphics[width=\textwidth]{ 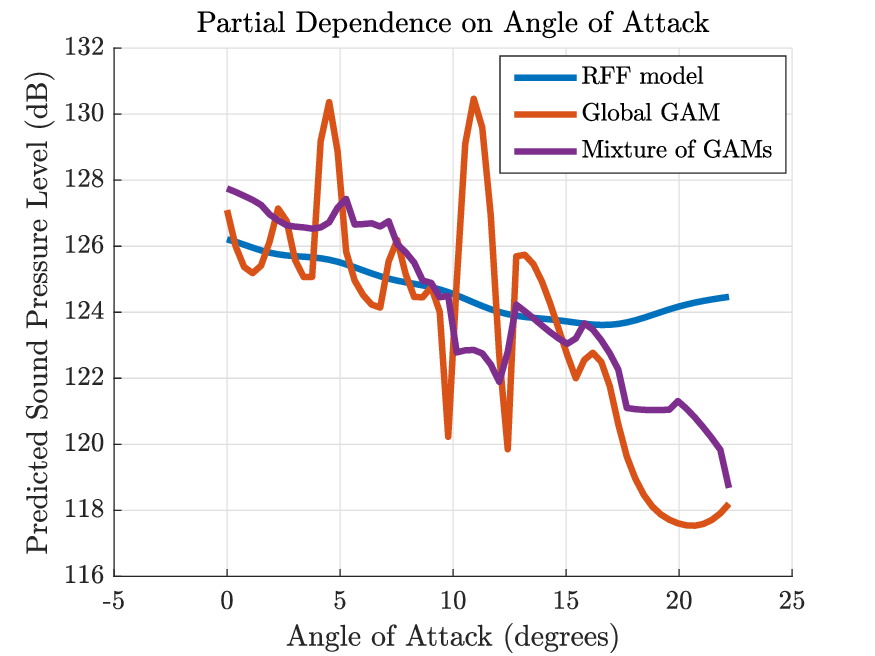}
    \caption{Angle of Attack}
\end{subfigure}
\hfill
\begin{subfigure}[b]{0.32\textwidth}
    \includegraphics[width=\textwidth]{ 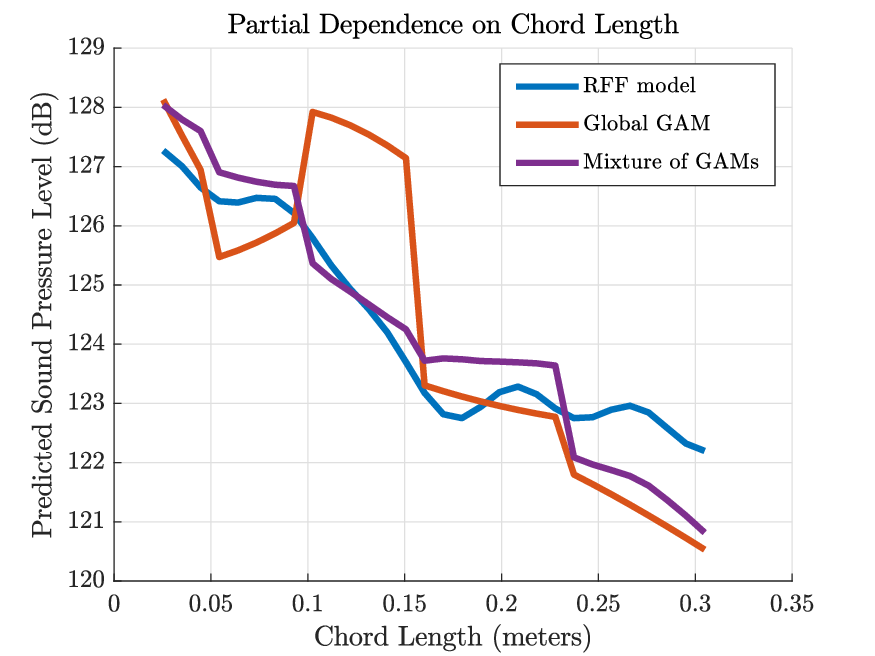}
    \caption{Chord Length}
\end{subfigure}

\vspace{1mm} 

\par\bigskip
\centering
\begin{subfigure}[b]{0.32\textwidth}
    \includegraphics[width=\textwidth]{ 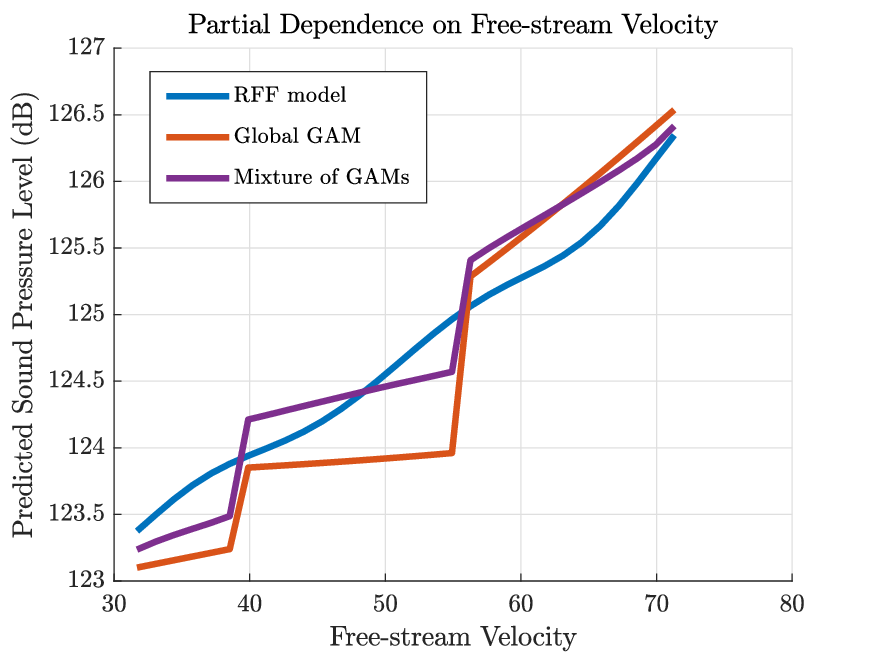}
    \caption{Free-stream Velocity}
\end{subfigure}
\hspace{0.05\textwidth}
\begin{subfigure}[b]{0.32\textwidth}
    \includegraphics[width=\textwidth]{ 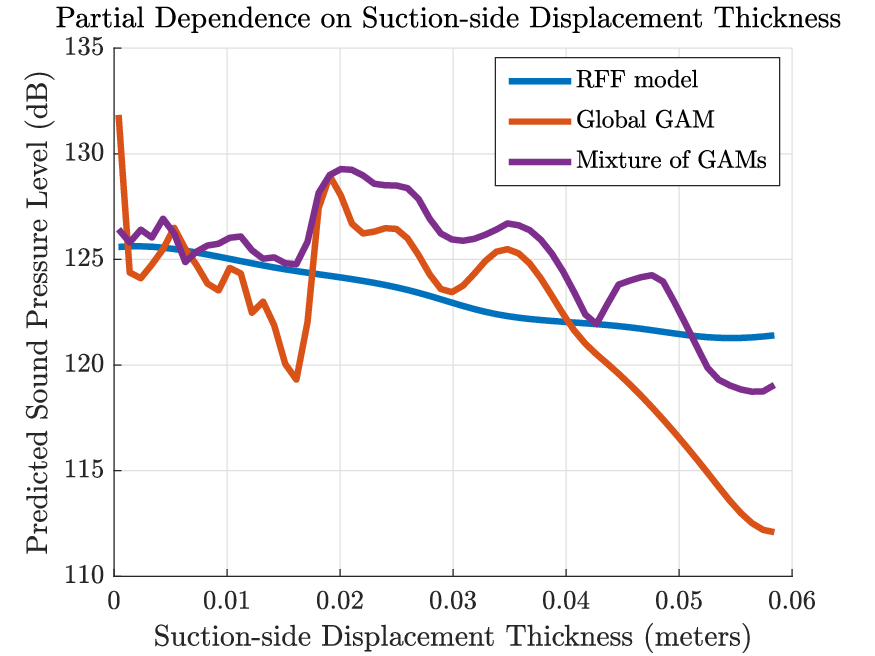}
    \caption{Displacement Thickness}
\end{subfigure}

\caption{Partial dependence plots on the five covariates of the Airfoil Self-noise dataset.}
\label{fig:pdp_airfoil}
\end{figure}

The validity of this local specialization is further corroborated by the partial dependence analysis in Figure~\ref{fig:pdp_airfoil}. Across several important variables, including acoustic frequency, angle of attack, and free-stream velocity, the covariate-response patterns produced by the mixture of GAMs (violet curves) more faithfully recover the nonlinear patterns of the RFF reference (blue curves) than the global GAM (red curves). This alignment indicates that the mixture structure allows the additive model to adapt to distinct regions of the aerodynamic response surface, rather than describing these coupled physical relationships as a single global effect. Consequently, the framework provides a more accurate interpretable approximation of the structural complexities identified by the underlying spectral model.

\subsubsection{Component ablation: representation and local model class}

The preceding case studies suggest that the proposed framework benefits from two complementary design choices: constructing regimes from a response-informed RFF representation and fitting nonlinear additive models within the resulting regimes. To further disentangle the contributions of these core components, we conduct an ablation study on the California Housing and Airfoil Self-Noise datasets. The variants reported in Table~\ref{tab:ablation_clustering_local_models} focus on two factors: the clustering representation and the local predictive model class.

The first two variants perform clustering either on the original input features or on a PCA-reduced representation of the raw covariates, followed in both cases by local GAM fitting. These baselines assess whether the RFF-induced representation provides a more informative basis for regime discovery than representations derived directly from the covariate space. The third variant retains the RFF-induced clustering structure but replaces the local GAMs with $\ell_2$-regularized linear models under the same clustering assignments, thereby isolating the contribution of nonlinear additive local modeling. The final row corresponds to the full proposed framework.

\begin{table}[htbp]
\centering
\caption{Ablation study comparing clustering sources and local model expressiveness.
All methods use the same number of clusters and identical training protocols.
Lower RMSE indicates better predictive accuracy.}
\label{tab:ablation_clustering_local_models}
\begin{tabular}{l cc cc}
\toprule
& \multicolumn{2}{c}{California Housing}
& \multicolumn{2}{c}{Airfoil Self-Noise} \\
\cmidrule(lr){2-3} \cmidrule(lr){4-5}
Method
& Train RMSE & Test RMSE
& Train RMSE & Test RMSE \\
\midrule
Raw data + Local GAM
& 0.479 & 0.538 $\pm$ 0.023
& 2.90 & 3.35 $\pm$ 0.08 \\
PCA data + Local GAM
& 0.507 & 0.556 $\pm$ 0.023
& 2.50 & 3.01 $\pm$ 0.07 \\
RFF + Local Linear Model
& 0.629 & 0.627 $\pm$ 0.025
& 2.61 & 2.51 $\pm$ 0.05 \\
RFF-based Mixture of GAMs
& 0.442 & \textbf{0.489} $\pm$ 0.021
& 2.01 & \textbf{2.22} $\pm$ 0.05 \\
\bottomrule
\end{tabular}
\end{table}

Across both datasets, the full RFF-based mixture of GAMs achieves the lowest test RMSE. Comparisons with the raw-data and PCA-data variants show that clustering in the spectral representation-guided latent space is more effective than clustering based on input-space alternatives. The comparison with the local linear variant further shows that, even under the same RFF-induced regimes, nonlinear additive local models provide a substantial improvement over linear experts. These results confirm that both the representation used for regime discovery and the expressiveness of the local model class contribute meaningfully to the performance of the proposed framework.

\section{Conclusion}\label{section_conclusion}
This work introduces a locally adaptive regression framework that combines random Fourier feature representations with cluster-specific generalized additive models. The method first uses an RFF regression model to construct a response-informed spectral feature map. This representation is then compressed by principal component analysis and partitioned by a Gaussian mixture model to identify latent predictive regimes. Within each regime, a GAM models nonlinear covariate effects through smooth univariate components, and the final predictor is formed as a soft mixture of these local additive models.

Numerical experiments on six benchmark regression datasets, including California Housing, NASA Airfoil Self-Noise, Bike Sharing, Kin40k, Elevators, and Protein, demonstrate the effectiveness of the proposed mixture-of-GAMs framework. Across all datasets, the method improves upon global interpretable baselines such as LASSO, MARS, and global GAM, highlighting the benefit of combining additive structure with local adaptivity. On California Housing and Bike Sharing, the proposed method also attains lower test errors than mixture-of-linear-model baselines, suggesting that smooth nonlinear local components can provide advantages over purely linear local models. More broadly, the results indicate that the framework remains competitive across regression settings with varying degrees of nonlinearity, heterogeneity, and data complexity.

Beyond predictive accuracy, the framework provides an interpretable regime-level structure. In the California Housing dataset, spatial Fourier features reveal regimes aligned with dominant geographic variation, including an inland-coastal gradient. In the Bike Sharing dataset, the learned posterior responsibilities exhibit distinct temporal activation patterns, reflecting changes in demand across different hours of the day. These examples illustrate that the response-informed spectral representation can support both regime discovery and interpretation, especially when the underlying covariates have meaningful spatial or temporal structure.

Taken together, the results suggest that the proposed framework offers a practical compromise between predictive flexibility and model transparency. By using RFFs to guide regime construction and GAMs to model local covariate effects, the method provides a principled alternative to both global explainable models and locally linear mixture-based approaches. The empirical findings also show that local adaptivity can recover a substantial part of the predictive strength of more flexible models while retaining interpretable component-wise structure.

Several directions remain for future research. One promising extension is to spatio-temporal regression, where random Fourier features could jointly encode spatial variation and temporal dynamics. Such representations may enable interpretable localized models for evolving physical, environmental, or socio-economic systems. Further work could also explore alternative latent representations, multiscale or hierarchical regime structures, and clustering methods that incorporate kernel-induced similarity more directly.

\section*{Data Availability}
The data used in this study consist of publicly available benchmark datasets, including the California Housing, NASA Airfoil Self-Noise, Bike Sharing, Kin40k, Elevators, and Protein benchmarks, which are cited in the references. The code used to generate the results is available in a public GitHub repository, with the link provided in Appendix~\ref{app:baselines}.

\section*{CRediT authorship contribution statement}
\textbf{Xin Huang}: Writing - review \& editing, Writing - original draft, Visualization, Software, Methodology, Investigation, Validation, Formal analysis, Conceptualization. \textbf{Jia Li}: Writing - review \& editing, Methodology, Formal analysis, Conceptualization. \textbf{Jun Yu}: Writing - review \& editing, Methodology, Formal analysis, Conceptualization, Resources.


\section*{Acknowledgements}
This work was supported by Kempe Stiftelserna project JCSMK23-0168. The computations used resources provided by the National Academic Infrastructure for Supercomputing in Sweden (NAISS) at the PDC Center for High Performance Computing, KTH Royal Institute of Technology, which is  partially funded by the Swedish Research Council through grant agreement no. 2022-06725. Jia Li's research is supported by the National Science Foundation under grant CCF-2205004.

\appendix
\section{Implementation and Hyperparameter Details}
\label{app:implementation}

\subsection{Hyperparameters for the mixture-of-GAMs framework}
\label{app:mgam_hyperparams}

This subsection summarizes the hyperparameter choices used for training the proposed Mixture-of-GAMs framework, including the resampling-based RFF representation, latent-space clustering, and local GAM fitting. A detailed overview of the final hyperparameter configurations for all datasets is provided in Table~\ref{tab:hyperparams_summary}.

For each dataset, two key structural hyperparameters, including the number of mixture components $L$ and the dimension $d$ of the PCA-induced latent embedding, are determined via grid search over moderate ranges of integer values. Specifically, candidate values of $L$ and $d$ are chosen according to the dataset size and complexity, and for each pair $(L,d)$, the model is trained on the training data and evaluated using predictive performance. The configuration yielding the best performance is selected and then fixed for all reported experiments on that dataset.

\begin{figure}[htbp]
    \centering
    \begin{subfigure}[b]{0.49\textwidth}
    \includegraphics[width=0.9\linewidth]{  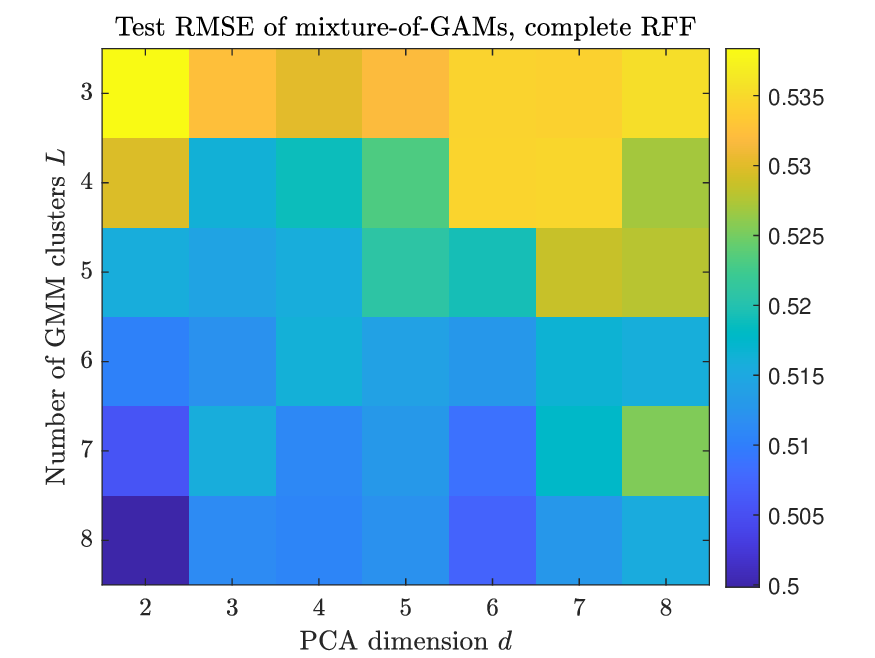}
     \caption{Clustering guided by complete RFF.}
    \end{subfigure}
    \begin{subfigure}[b]{0.49\textwidth}
    \includegraphics[width=0.9\linewidth]{ 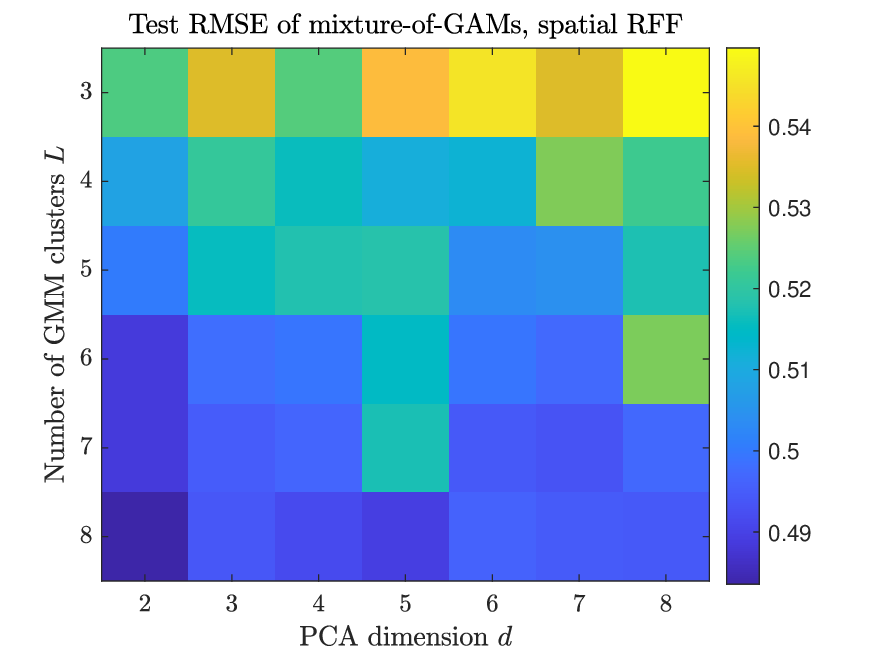}
     \caption{Clustering guided by spatial RFF.}
    \end{subfigure}
    \caption{ Test root mean square error on California housing dataset evaluated over a grid of hyperparameter configurations $(L, d)$, where $L$ is the number of mixture components and $d$ is the number of retained principal components after PCA on the intermediate feature representations. Panel (a) uses the full set of RFF features to guide the Gaussian mixture model-based clustering, while panel (b) relies solely on spatial RFF features derived from geographic coordinates.}
    \label{fig:RMSE_grid}
\end{figure}

\begin{figure}[htbp]
    \centering
    \begin{subfigure}[b]{0.49\textwidth}
    \includegraphics[width=0.9\linewidth]{ 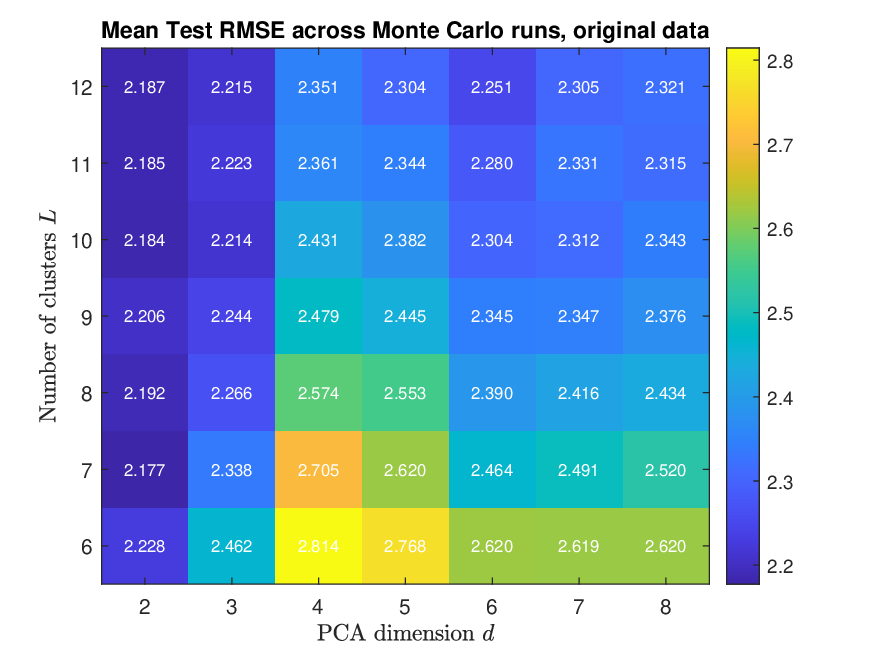}
     \caption{Original dataset.}
    \end{subfigure}
    \begin{subfigure}[b]{0.49\textwidth}
    \includegraphics[width=0.9\linewidth]{ 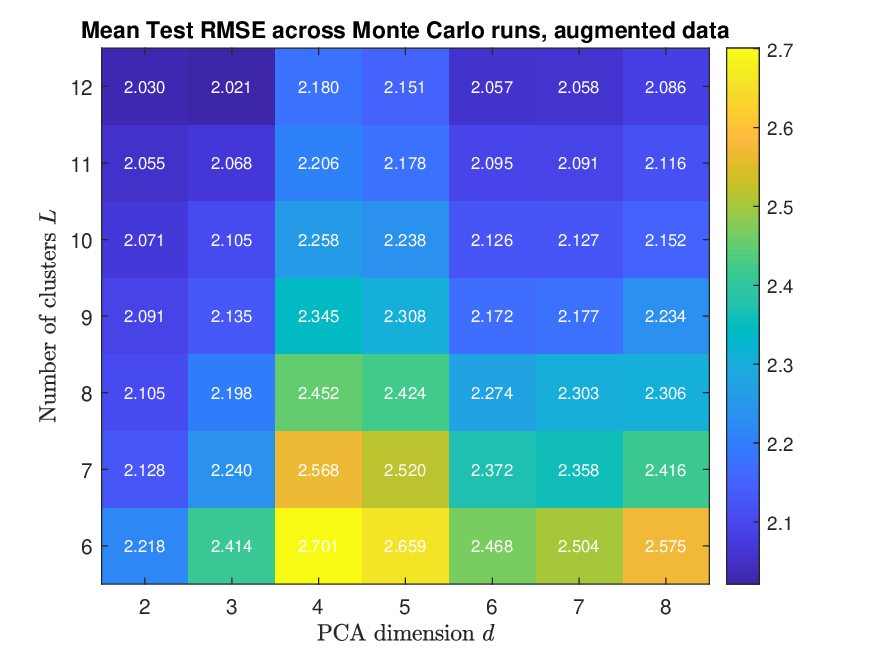}
     \caption{Augmented dataset.}
    \end{subfigure}
    \caption{Test root mean square error on the airfoil self-noise dataset evaluated over a grid of hyperparameter configurations $(L, d)$, where $L$ is the number of mixture components and $d$ is the number of retained PCA components. Panel (a) uses the original dataset, while panel (b) uses the RFF-augmented dataset.}
    \label{fig:RMSE_grid_airfoil}
\end{figure}

To illustrate the influence of these hyperparameters, we visualize the performance landscape over $(L,d)$ using heatmaps, where each grid cell represents the test RMSE associated with a given configuration. Figure~\ref{fig:RMSE_grid} displays the results for the California Housing dataset using both complete and spatial RFF representations, while Figure~\ref{fig:RMSE_grid_airfoil} presents the corresponding results for the Airfoil Self-Noise dataset under both original and augmented data settings.

Across these experiments, several consistent patterns can be observed. First, increasing the number of mixture components $L$ generally improves predictive accuracy, reflecting the benefit of increased local adaptivity. In contrast, the model's sensitivity to the latent dimension $d$ is relatively low, with robust performance achieved even at small values of $d$.

In addition, the choice of representation and data augmentation strategy has a clear impact on the performance landscape. For the California Housing dataset, clustering guided by spatial random Fourier features yields lower RMSE across a broad range of configurations, indicating that the spatial representation provides a more informative basis for regime discovery. For the Airfoil dataset, data augmentation consistently improves performance across the hyperparameter grid, suggesting that the additional samples stabilize the resulting cluster structures.

These observations support the use of moderate values of $L$ and low-dimensional latent embeddings in the proposed framework, and provide empirical justification for the hyperparameter choices reported in Table~\ref{tab:hyperparams_summary}.

For local GAM fitting, we use the fitting weights $a_{i\ell}$ defined in Section~\ref{sec:clusterwise_gam}. In the experiments, hard assignment weights are used for California Housing, Airfoil Self-Noise, and Protein, while soft posterior-responsibility weights are used for Bike Sharing, Kin40k, and Elevators. For the hard-assignment cases, local GAMs are fitted using a custom spline-based penalized least-squares implementation. For the soft-weighted cases, we interface MATLAB with the Python package \texttt{pygam} (Serv\'en and Brummitt, \citeyear{serven2018pygam}) and pass the GMM responsibilities as observation weights for each local GAM. In the Bike Sharing experiment, cyclic spline terms are used for temporal covariates, smooth spline terms for continuous meteorological variables, and one-hot encoded linear terms for categorical indicators.

\begin{table}[t]
\centering
\small
\caption{Summary of hyperparameters used for the resampling-based RFF model and the Mixture-of-GAMs framework across all six datasets.}
\label{tab:hyperparams_summary}
\begin{tabular}{l c c c}
\toprule
\textbf{Hyperparameter} & \textbf{Cal. Housing} & \textbf{Airfoil} & \textbf{Bike Sharing} \\
\midrule

\multicolumn{4}{l}{\textbf{RFF Settings}} \\
Number of Fourier features $K$  & 4000 & 2000 & 2000 \\
Random walk step size $\delta$  & 0.3 & 0.1 & 0.05 \\
Tikhonov regularization $\lambda$  & 0.32 & 0.06 & 0.04 \\

\addlinespace
\multicolumn{4}{l}{\textbf{Mixture-of-GAMs Settings}} \\
PCA dimension $d$  & 2 & 3 & 3 \\
Number of GMM clusters $L$  & 8 & 12 & 8 \\
Knots per feature  & 30 & 10 & 10 \\
Spline degree  & 3 & 3 & 3 \\
Smoothing penalty & 2nd diff. & 2nd diff. & 2nd diff. \\

\addlinespace
\multicolumn{4}{l}{\textbf{Training Setup}} \\
Train/test split  & 80/20 & 80/20 & 80/20 \\
Training set size $N$ & 16512 & 1202 & 13903 \\

\midrule
\multicolumn{4}{l}{\textbf{(continued) Remaining datasets}} \\
\midrule
\textbf{Hyperparameter} & \textbf{Kin40k} & \textbf{Elevators} & \textbf{Protein} \\
\midrule

\multicolumn{4}{l}{\textbf{RFF Settings}} \\
Number of Fourier features $K$  & 4000 & 1000 & 12000 \\
Random walk step size $\delta$  & 0.1 & 0.1 & 0.2 \\
Tikhonov regularization $\lambda$  & 0.02 & 0.08 & 0.06 \\

\addlinespace
\multicolumn{4}{l}{\textbf{Mixture-of-GAMs Settings}} \\
PCA dimension $d$  & 8 & 3 & 8 \\
Number of GMM clusters $L$  & 46 & 12 & 22 \\
Knots per feature  & 10 & 10 & 10 \\
Spline degree  & 3 & 3 & 3 \\
Smoothing penalty & 2nd diff. & 2nd diff. & 2nd diff. \\

\addlinespace
\multicolumn{4}{l}{\textbf{Training Setup}} \\
Train/test split  & 90/10 & 90/10 & 90/10 \\
Training set size $N$ & 36000 & 14940 & 41157 \\

\bottomrule
\end{tabular}
\end{table}

\subsection{Baseline models and training configuration}
\label{app:baselines}

We compare the proposed Mixture-of-GAMs framework with a set of commonly used baseline models for regression, using standard and widely adopted software packages. Random Forest regression was implemented using the \texttt{RandomForestRegressor} class from the \texttt{scikit-learn} library (Pedregosa et al. \citeyear{scikit_learn_jmlr}), with 200 trees and square-root feature subsampling at each split. Multivariate adaptive regression splines (MARS) were trained using the \texttt{Earth} implementation from the \texttt{pyearth} package, employing cubic spline basis functions and default knot selection. Gradient boosting models were implemented using the \texttt{xgboost} package, with the number of boosting iterations fixed to 5000 across all four datasets. Explainable Boosting Machine (EBM) models were trained with a limited number of interaction terms, using 5 to 8 pairwise interactions in total to control model complexity. For these baseline implementations, the hyperparameters were selected based on five-fold cross-validation on the training data. For the NASA Airfoil Self-Noise dataset, the acoustic frequency variable was log-transformed prior to model fitting. 

The multilayer perceptron (MLP) and its associated mixture-of-linear-model variants (MLM-cell and MLM-epic) were trained using the same hyperparameter settings as in Seo et al. (\citeyear{MLM}), ensuring comparability across model classes. Full implementation details, including training scripts and configuration files for all baseline models, are provided in the accompanying public code repository:
\\\href{https://github.com/XinHuang2022/Mixture_of_GAMs_Informed_by_Fourier_Features}{https://github.com/XinHuang2022/Mixture\_of\_GAMs\_Informed\_by\_Fourier\_Features}.

\section{Nomenclature and Abbreviations}
\label{app:notations}

In this section, we summarize the abbreviations and notation used in this work, as listed in Tables~\ref{tab:abbreviations} and~\ref{tab:nomenclature_new}.

\begin{table}[h!]
\centering
\small
\caption{Abbreviations}
\label{tab:abbreviations}
\begin{tabular}{ p{3.5cm} p{10.5cm} }
\toprule
\textbf{Abbreviation} & \textbf{Description} \\
\midrule
RFF & Random Fourier Features \\
GAM & Generalized Additive Model \\
DNN & Deep Neural Network \\
GMM & Gaussian Mixture Model \\
MLM & Mixture of Linear Models \\
PCA & Principal Component Analysis \\
SVD & Singular Value Decomposition \\
EM & Expectation-Maximization (algorithm) \\
LASSO & Least Absolute Shrinkage and Selection Operator \\
MLP & Multi-Layer Perceptron \\
MARS & Multivariate Adaptive Regression Splines \\
RBF & Radial Basis Function \\
\bottomrule
\end{tabular}
\end{table}

\begin{table}[h!]
\centering
\small
\caption{Nomenclature}
\label{tab:nomenclature_new}
\begin{tabular}{ p{3.8cm} p{10.2cm} }
\toprule

\multicolumn{2}{l}{\textbf{Vectors, Sets, and Data}} \\
\midrule
$\{(\bmx_i,y_i)\}_{i=1}^N$ 
& Training dataset consisting of $N$ input-output pairs \\

$\bmx=[x_1,\dots,x_p]^\top \in \mathbb{R}^p$ 
& Input vector with $p$ covariates \\

$y_i \in \mathbb{R}$ 
& Response variable associated with input $\bmx_i$ \\


$\{\bmomega_k\}_{k=1}^K$ with $\bmomega_k \in \mathbb{R}^p$
& Set of sampled random Fourier frequencies \\

$\varsigma(\bmx) \in \mathbb{C}^K$ 
& Random Fourier feature map evaluated at $\bmx$ \\

$\bm{\beta} \in \mathbb{C}^K$ 
& Coefficient vector of the random Fourier feature model \\

\midrule
\multicolumn{2}{l}{\textbf{Feature Matrices and PCA Components}} \\
\midrule
$\mathbf{S} \in \mathbb{R}^{N\times K}$ 
& Random Fourier feature matrix for the training data \\

$\bar{\mathbf{S}}$ 
& Centered random Fourier feature matrix \\

$\bar{\mathbf{s}} \in \mathbb{R}^K$ 
& Empirical mean of the random Fourier feature vectors \\

$\mathbf{U},\,\mathbf{\Sigma},\,\mathbf{V}$ 
& Matrices from the singular value decomposition of $\bar{\mathbf{S}}$ \\

$\mathbf{V}_d \in \mathbb{R}^{d\times K}$ 
& PCA projection matrix retaining the leading $d$ principal components \\

$\mathbf{Z} \in \mathbb{R}^{N\times d}$ 
& Low-dimensional latent embedding of the training data \\

\midrule
\multicolumn{2}{l}{\textbf{Functions and Predictive Models}} \\
\midrule
$m(\bmx)$ 
& Target regression function \\

$\bar{m}(\bmx)$ 
& Random Fourier feature approximation of $m(\bmx)$ \\

$h(\bmx)$ 
& PCA-reduced latent representation of input $\bmx$ \\

$\tilde f^{(\ell)}(\bmx)$ 
& Cluster-specific generalized additive model \\

$g_j^{(\ell)}(x_j)$ 
& Univariate smooth function for covariate $j$ in cluster $\ell$ \\

$\phi_{j,q}(x_j)$ 
& Spline basis function utilized in the generalized additive model \\

$\tilde m(\bmx)$ 
& Aggregated mixture-of-GAMs regression function \\

\midrule
\multicolumn{2}{l}{\textbf{Mixture Model Parameters}} \\
\midrule
$L$ 
& Number of mixture components (clusters) \\

$\pi_\ell$ 
& Mixing weight of the $\ell$-th Gaussian component \\

$\mu_\ell,\;\Sigma_\ell$ 
& Mean and covariance of the $\ell$-th Gaussian component \\

$\gamma_\ell\big(h(\bmx)\big)$ 
& Posterior responsibility of cluster $\ell$ for input $\bmx$ \\

$a_{i\ell}$ 
& Fitting weight of the $i$-th observation for training the $\ell$-th local GAM \\

\midrule
\multicolumn{2}{l}{\textbf{Miscellaneous Symbols}} \\
\midrule
$\odot$ 
& Hadamard (elementwise) product \\

$\mathrm{Re}(\cdot)$ 
& Real part of a complex-valued quantity \\

$\mathcal{N}(\cdot;\mu,\Sigma)$ 
& Multivariate Gaussian distribution with mean $\mu$ and covariance $\Sigma$ \\

\bottomrule
\end{tabular}
\end{table}

\end{document}